\documentclass[10pt, letter, onecolumn]{arxiv}

\usepackage{kantlipsum, lipsum}
\usepackage{dm-colors}
\usepackage{amsmath}
\usepackage{pstricks, pst-node}
\usepackage{verbatim}
\usepackage{multirow}
\usepackage{scalerel}
\usepackage{booktabs}
\usepackage{enumitem}
\usepackage{xspace}
\usepackage{bm}
\usepackage{bbm}
\usepackage{mathtools}
\usepackage{soul}
\usepackage{epsfig}
\usepackage{graphicx}
\usepackage{amssymb}
\usepackage{colortbl}
\usepackage{csquotes}
\usepackage{setspace}
\usepackage{colortbl}
\usepackage{tabularx,ragged2e}
\usepackage{placeins}
\usepackage[symbol]{footmisc}
\usepackage{nameref}
\usepackage{varioref}
\usepackage[pagebackref=false,breaklinks=false,%
            colorlinks=true,bookmarks=true,citecolor=ourdarkblue,%
            urlcolor=ourdarkblue,linkcolor=ourdarkblue]{hyperref}
\usepackage[capitalize]{cleveref}
\usepackage{etoc}

\graphicspath{{figures/}}

\definecolor{purple}{RGB}{100,0,200}

\title{\Large{Towards Generic Anomaly Detection and Understanding:\\ Large-scale Visual-linguistic Model (GPT-4V) Takes the Lead} }

\vspace{1cm}

\author[1$\ast$]{Yunkang Cao} 
\author[2$\ast$]{Xiaohao Xu}
\author[3$\ast$]{Chen Sun}
\author[2]{Xiaonan Huang}
\author[1]{Weiming Shen}

\affil[1]{\normalsize Huazhong University of Science and Technology \hspace{8pt}}

\affil[2]{\normalsize University of Michigan, Ann Arbor \authorcr \vspace{3pt} 
}

\affil[3]{\normalsize University of Toronto \authorcr}

\renewcommand{\correspondingauthor}[1]{$\ast$~Authors contribute equally. Email: cyk\_hust@hust.edu.cn, xiaohaox@umich.edu, chrn.sun@mail.utoronto.ca, 
 xiaonanh@umich.edu, wshen@ieee.org}

\definecolor{HLYellow}{RGB}{255,255,0}
\definecolor{HLRed}{RGB}{180,0,0}
\definecolor{HLGreen}{RGB}{112,173,71}
\definecolor{HLBlue}{RGB}{68,114,196}

\sethlcolor{HLYellow}

\begin{document}

\begin{abstract}
\section*{\centering Abstract}
Anomaly detection is a crucial task across different domains and data types. However, existing anomaly detection models are often designed for specific domains and modalities. This study explores the use of GPT-4V(ision), a powerful visual-linguistic model, to address anomaly detection tasks in a generic manner. We investigate the application of GPT-4V in multi-modality, multi-domain anomaly detection tasks, including image, video, point cloud, and time series data, across multiple application areas, such as industrial, medical, logical, video, 3D anomaly detection, and localization tasks. To enhance GPT-4V's performance, we incorporate different kinds of additional cues such as class information, human expertise, and reference images as prompts. Based on our experiments, GPT-4V proves to be highly effective in detecting and explaining global and fine-grained semantic patterns in zero/one-shot anomaly detection. This enables accurate differentiation between normal and abnormal instances. Although we conducted extensive evaluations in this study, there is still room for future evaluation to further exploit GPT-4V's generic anomaly detection capacity from different aspects. These include exploring quantitative metrics, expanding evaluation benchmarks, incorporating multi-round interactions, and incorporating human feedback loops. Nevertheless, GPT-4V exhibits promising performance in generic anomaly detection and understanding, thus opening up a new avenue for anomaly detection.

\vspace{0.2cm}
All evaluation samples, including image and text prompts, will be available at \url{https://github.com/caoyunkang/GPT4V-for-Generic-Anomaly-Detection}.

\end{abstract}

\maketitle


\vspace{1cm}
\tableofcontents
\newpage

\listoffigures

\newpage

\section{Introduction}

\subsection{Motivation and Overview}

Anomaly detection~\cite{AD_SURVEY, Chalapathy2019DeepLF,Pang2020DeepLF,BlazquezGarcia2020ARO,ruff_unifying_2021} involves identifying data patterns or data points that significantly deviate from normality. These anomalies or outliers are rare, unusual, or inconsistent data points that deviate from the majority of the data. The primary objective of anomaly detection is to automatically detect and pinpoint these irregularities, which may signify errors, fraud, unusual events, or other noteworthy phenomena, facilitating further investigation or necessary action. Anomaly detection techniques have been widely employed in diverse domains, such as industrial inspection~\cite{diers_survey_2023,IM-IAD}, medical diagonisis~\cite{Zhang2020COVID19SO}, video surveillance~\cite{Sultani2018RealWorldAD}, fraud detection~\cite{Du2017DeepLogAD} and many other areas where identifying unusual instances is crucial.

Despite the existence of numerous techniques~\cite{IKD, GANomaly, vijay2010, huang_weakly_2022, hasan2016learning, sakurada2014anomaly, zhao2020multivariate, CPMF, Yao2023LGC} for anomaly detection, many existing approaches predominantly rely on methods that describe the normal data distribution. They often overlook high-level perception and primarily treat it as a low-level task. However, practical applications of anomaly detection frequently necessitate a more comprehensive, high-level understanding of the data. Achieving this understanding entails at least three crucial steps:

\begin{enumerate}
    \item \textbf{Understanding the Data Types and Categories:} The first step involves a thorough comprehension of the data types and categories present in the dataset. Data can take various forms, including images, videos, point clouds, time-series data, etc. Each data type may require specific methods and considerations for anomaly detection. Furthermore, different categories may have distinct definitions of normal states.
    \item \textbf{Determining Standards for Normal States:} After obtaining the data types and categories, it would be feasible to further reason the standards for normal states, which requires a high-level understanding of the data.
    \item \textbf{Evaluating Data Conformance:} The final step is to assess whether the provided data conforms to the established standards for normality. Any deviation from these standards can be categorized as an anomaly.
\end{enumerate}

Recent advancements in large multimodal models (LMMs)~\cite{chowdhery2022palm,anil2023palm,gong2023multimodalgpt,zhu2023minigpt,liu2023visual,dai2023instructblip,li2023multimodal} have shown robust reasoning capacity~\cite{liu2023hallusionbench,liu2023covid} and created new opportunities for improving anomaly detection. LMMs are typically trained on extensive multimodal datasets~\cite{schuhmann2022laion}, enabling them to effectively analyze various data types, including natural language and visual information. They hold the potential to address the challenges associated with high-level anomaly detection~\cite{AnomalyGPT,SAA,VAND,AnomalyCLIP}. 

Moreover, OpenAI recently introduced GPT-4V(ision)~\cite{Dawn-of-LMMs}, a state-of-the-art LMM that has exhibited remarkable performance across various practical applications. However, it remains uncertain whether GPT-4V can also exhibit robust capabilities for anomaly detection. The objective of this study is to bridge this knowledge gap by assessing the anomaly detection capabilities of GPT-4V.

\begin{figure}[hbt!]
    \centering
    \includegraphics[width = \textwidth]{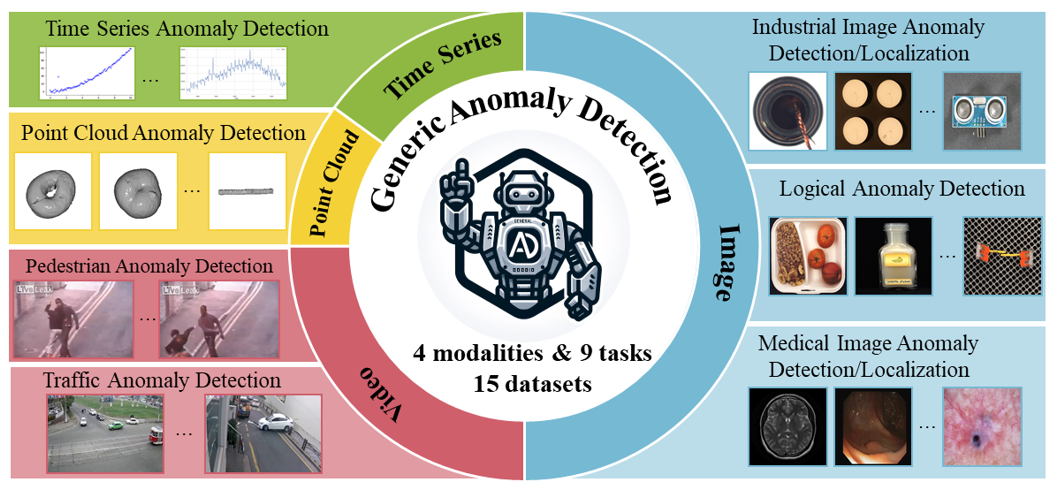}
    \vspace{3pt}
    \caption[The Diagram of Evaluation GPT-4V on Multi-modality/fields Anomaly Detection.]
    {\textbf{Comprehensive Evaluation of GPT-4V for Multi-modality Multi-task Anomaly Detection}
In this study, we conduct a thorough evaluation of GPT-4V in the context of multi-modality anomaly detection. We consider four modalities: image, video, point cloud, and time series, and explore nine specific tasks, including industrial image anomaly detection/localization, point cloud anomaly detection, medical image anomaly detection/localization, logical anomaly detection, pedestrian anomaly detection, traffic anomaly detection, and time series anomaly detection. Our evaluation encompasses a diverse range of 15 datasets.
}
    \vspace{3pt}
    \label{fig:framework}
\end{figure}

\subsection{Our Approach: Prompting GPT-4V for Anomaly Detection}

\subsubsection{Prompt Designs}

The design of prompts plays a crucial role in effectively directing GPT-4V's attention toward the specific aspects of the anomaly detection task. In this study, we primarily consider four types of prompts:

\begin{enumerate}
    \item \textbf{Task Information Prompt}:  To prompt GPT-4V effectively for anomaly detection, it is essential to provide clear task information. This study formulates the prompt as follows: "Please determine whether the image contains anomalies or outlier points."
    \item \textbf{Class Information Prompt}: The understanding of data types and categories is critical. In cases where GPT-4V may struggle to recognize the data class, explicit class information may be provided. For instance, "Please determine whether the image, which is related to the \{CLS\}, contains anomalies or defects."
    \item \textbf{Normal Standard Prompt}: GPT-4V may encounter difficulties in answering questions related to determining normal standards, and sometimes the standards even can not be examined without human expertise. Hence, this study also explicitly provides the normal standards. For example, normal standards for the breakfast box in MVTec-LOCO~\cite{MVTec-LOCO} could be expressed as follows: "1. It should contain two oranges, one peach, and some cereal, nuts, and banana slices; 2. The fruit should be on the left side of the lunchbox, the cereal on the upper right, and the nuts and banana slices on the lower right of the lunchbox."
    \item \textbf{Reference Image Prompt}: To ensure better alignment between normal standards and images, a normal reference image is provided alongside language prompts. For example, "The first image is normal. Please determine whether the second image contains anomalies or defects."
\end{enumerate}

The study aims to explore how the use of these prompts, either individually or in different combinations depending on certain cases, impacts GPT-4V's capacity for anomaly detection.

\subsubsection{Evaluation Scope: Modalities and Domains}

Extensive evaluations are conducted in this study to assess the capabilities of GPT-4V in anomaly detection, as Fig.~\ref{fig:framework} shows. From the perspective of modalities, we evaluate image (Section \ref{Industrial Image Anomaly Detection}, \ref{Industrial Image Anomaly Localization}, \ref{Logical Anomaly Detection}, \ref{Medical Image Anomaly Detection}, \ref{Medical Image Anomaly Localization}), 
point cloud (Section \ref{Point Cloud Anomaly Detection}), 
video (Section \ref{Pedestrian Anomaly Detection}, \ref{Traffic Anomaly Detection}), 
and time series (Section \ref{Time Series Anomaly Detection}). From the perspective of fields, industrial inspection (Section \ref{Industrial Image Anomaly Detection}, \ref{Industrial Image Anomaly Localization}, \ref{Logical Anomaly Detection}, \ref{Point Cloud Anomaly Detection}), medical diagnosis (Section \ref{Medical Image Anomaly Detection}, \ref{Medical Image Anomaly Localization}), and video surveillance (Section \ref{Pedestrian Anomaly Detection}, \ref{Traffic Anomaly Detection}) are evaluated. To the best of our knowledge, this is the first study to investigate such a wide range of modalities and fields for anomaly detection.

\subsection{Limitations in Anomaly Detection Evaluation Based on GPT-4V}

The analysis of this study is subject to certain limitations:
\begin{enumerate}
    \item \textbf{Predominance of Qualitative Results}:
    The analysis primarily relies on qualitative assessment, lacking quantitative metrics that could offer a more objective evaluation of the model's performance in anomaly detection. Incorporating quantitative measures would provide a more robust basis for assessment.
    
    \item \textbf{Scope of Evaluated Cases}:
    The evaluation is confined to a limited scope of cases or scenarios. This narrow focus may not fully capture the diverse challenges encountered in real-world anomaly detection tasks. Expanding the range of evaluated cases would yield a more comprehensive understanding of the model's capabilities.
    
    \item \textbf{Single Interaction Evaluation}:
    The study mainly concentrates on a single-round conversation. In contrast, multi-round conversations, as observed in the in-context learning capacity of GPT-4V~\cite{Dawn-of-LMMs}, can stimulate deeper interaction. The single-round conversation approach restricts the depth of interaction and may constrain the model's comprehension and its effectiveness in responding to anomaly detection tasks. Exploring multi-round interactions could reveal a more nuanced perspective of the model's performance.
\end{enumerate}

\section{Observations of GPT-4V on Multi-modal Multi-domain Anomaly Detection }

Following a thorough evaluation of GPT-4V's performance across various multi-modality and multi-field anomaly detection tasks, it becomes apparent that GPT-4V possesses robust anomaly detection capabilities. More precisely, GPT-4V consistently excels in addressing the three previously mentioned challenges: comprehending image context, discerning normal standards, and effectively comparing the provided image against these standards. In addition to these fundamental findings, our assessments have yielded valuable insights.

\subsection{GPT-4V can address multi-modality and multi-field anomaly detection tasks in zero/one-shot regime:}
\textbf{Anomaly detection for multi-modality}: GPT-4V's ability to handle diverse data modalities is demonstrated by its consistent performance across various domains. For instance, it exhibits proficiency in identifying anomalies in images, point clouds, X-rays, etc., underscoring its adaptability to multi-modal tasks. This versatility allows it to transcend the limitations of single-modal anomaly detectors.

\textbf{Anomaly detection for multi-field}: GPT-4V's performance across multiple fields, including industrial, medical, pedestrian, traffic, and time series anomaly detection, showcases its ability to seamlessly adapt to the distinct characteristics of each domain. Its consistent results affirm its broad applicability and versatility, making it a valuable tool for anomaly detection in a variety of real-world contexts.

\textbf{Anomaly detection in zero/one-shot regime}: GPT-4V's evaluation in both zero-shot and one-shot settings highlights its adaptability to different inference scenarios. In the absence of reference images, the model effectively relies on language prompts to detect anomalies. However, when provided with normal reference images, its anomaly detection accuracy is further enhanced. This flexibility enables GPT-4V to cater to a wide range of anomaly detection applications, whether with or without prior knowledge.

\subsection{GPT-4V can understand both global and fine-grained semantics for anomaly detection:}

\textbf{GPT-4V's understanding of global semantics}: GPT-4V's capacity to comprehend global semantics is demonstrated in its ability to recognize overarching abnormal patterns or behaviors. For example, in traffic anomaly detection, it can discern the distinction between typical traffic flow and irregular events, providing a holistic interpretation of the data. This global understanding makes it well-suited for identifying anomalies that deviate from expected norms in a broader context.

\textbf{GPT-4V's understanding of fine-grained semantics}: GPT-4V's fine-grained anomaly detection capabilities shine in cases where it not only detects anomalies but also precisely localizes them within complex data. For instance, in industrial image anomaly detection, it can pinpoint intricate details like slightly tilted wicks on candles or minor scratches or residues around the top rim of the bottle. This fine-grained understanding enhances its ability to detect subtle anomalies within complex data, contributing to its overall effectiveness.

\subsection{GPT-4V can automatically reason for anomaly detection:}
The model's strength in automatically reasoning the given complex normal standards and generating explanations for detected anomalies is a valuable feature. In logical anomaly detection, for example, GPT-4V excels at dissecting complex rules and providing detailed analyses of why an image deviates from the expected standards. This inherent reasoning ability adds a layer of interpretability to its anomaly detection results, making it a valuable tool for understanding and addressing irregularities in various domains.

\subsection{GPT-4V can be enhanced with increasing prompts:}
The results of the evaluation highlight the positive impact of additional prompts on GPT-4V's anomaly detection performance. The model's response to class information, human expertise, and reference images suggests that providing it with more context and information significantly improves its ability to detect anomalies accurately. This feature allows users to fine-tune and enhance the model's performance by providing relevant and supplementary information.

\subsection{GPT-4V can be constrained in real-world application but still promising:}
From the cases we test, we find there are still several gaps for GPT4V models to be applied in real world anomaly detection. For example, GPT-4V may face challenges in handling highly complex scenarios for industrial application. Ethical constraints in the medical field also make it conservative and hesitate to give confident answer. But we believe it remains promising in a wide range of anomaly detection tasks. To address these challenges effectively, further enhancements, specialized fine-tuning, or complementary techniques may be required. GPT-4V's potential for anomaly detection is evident, and ongoing research may continue to unlock its capabilities in even more complex scenarios.

\section{Industrial Image Anomaly Detection}
\label{Industrial Image Anomaly Detection}

\subsection{Task Introduction}

Industrial image anomaly detection is a critical component of manufacturing processes aimed at upholding product quality~\cite{MVTec-AD, IM-IAD, IKD}. Following the establishment of the MVTec AD dataset~\cite{MVTec-AD}, various methods~\cite{CDO, WinClip, VAND, SAA, CDO, MRKD,wan_logit_2022} have thrived in this field. These methods focus on determining whether testing images contain anomalies, typically represented as local structural variants. Early methods~\cite{PEFM, Student-Teacher, 2023YAODAT, SSKD, ocrgan, DFM} concentrated on developing specific models for given categories, while recent approaches~\cite{WinClip,VAND,SAA,AnomalyCLIP} target a more general but challenging solution, i.e., developing a unified model for arbitrary product categories, which usually performs in few-shot~\cite{Graphcore, huang2022registration} or even zero-shot~\cite{WinClip,SAA,VAND} regime. As highlighted in~\cite{Dawn-of-LMMs}, GPT-4V, equipped with extensive world knowledge, presents a promising solution for arbitrary category inspection.

\subsection{Testing philosophy}
Different prompts~\cite{Dawn-of-LMMs,liu2023aligning} could lead to different responses from GPT-4V. We aim to investigate the influence of different information on prompting GPT-4V for industrial anomaly detection. Following the previously discussed problems, this study further develops three prompts, a) class information: the names of the desired inspecting products, such as "bottle" and "candle", b) human expertise: the normal appearance and potential abnormal states and express them in languages, e.g., "Normally, the image given should show a clean and well-structured printed circuit board (PCB) with clear traces, soldered components, and distinct labels. It may have defects such as bent pins, cold solder joints, missing components, or smudged labels", c) reference image: normal reference image to provide GPT-4V a better understanding of normality. We propose to evaluate GPT-4V in either a zero-shot setting, with only language prompts, or a one-shot setting, with one reference image provided along with the language prompts. For each setting, we test three different variants: a) a naive prompt like "Please determine whether the image contains anomalies or defects," b) with class information, and c) with human expertise.

\subsection{Case Demonstration}

\begin{figure}[hbt!]
    \centering
    \includegraphics[width = \textwidth]{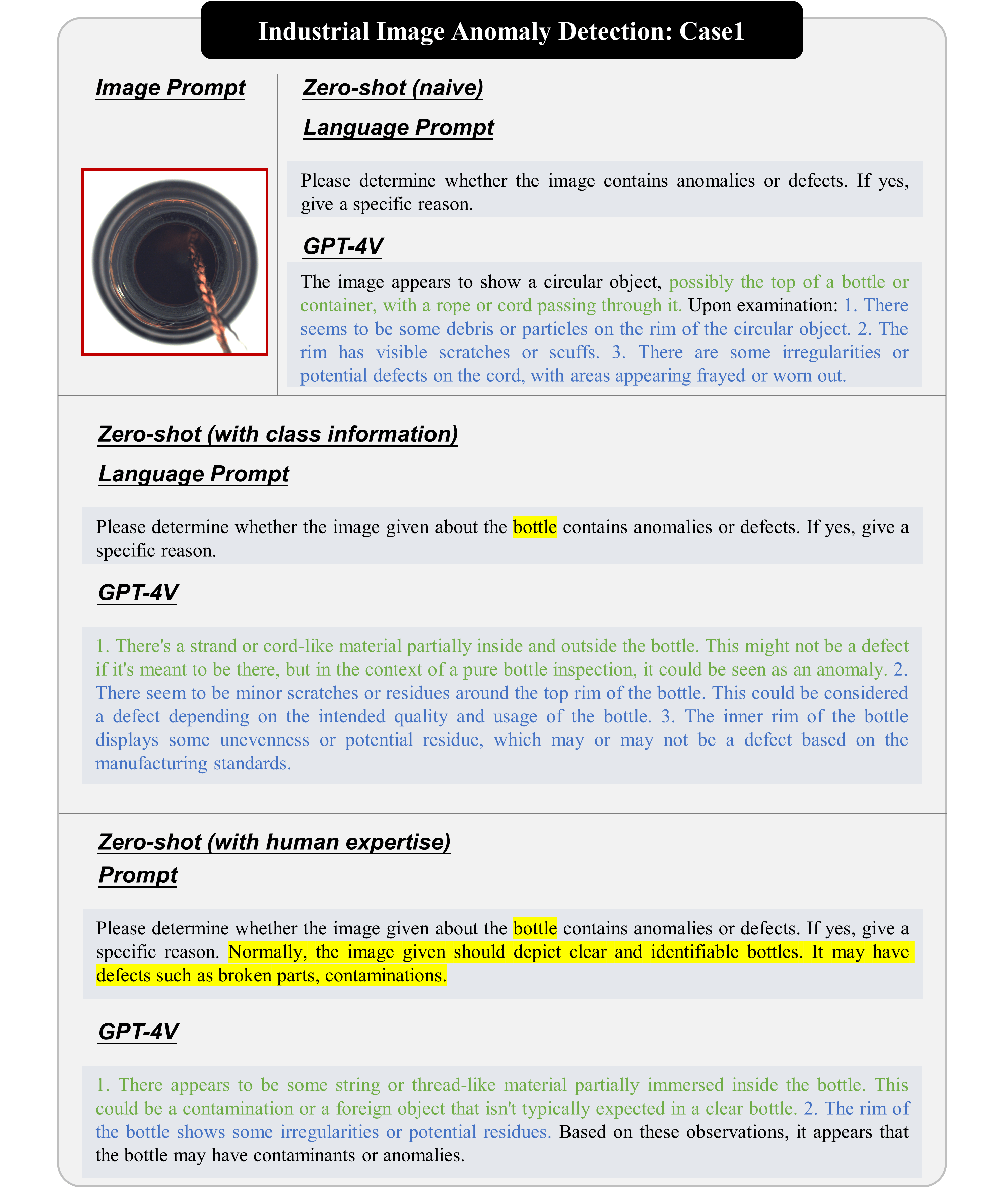}
    \vspace{3pt}
    \caption[Industrial Image Anomaly Detection: Case 1]
    {\textbf{Industrial Image Anomaly Detection: Case 1, zero-shot, the Bottle category of MVTec AD~\cite{MVTec-AD}}. \hl{Yellow} highlights the given class information and normal and abnormal state descriptions.   \textcolor{HLGreen}{Green}, \textcolor{HLRed}{red}, and \textcolor{HLBlue}{blue} highlight the expected, incorrect, and additional information outputted by GPT-4V. 
}
    \vspace{3pt}
    \label{fig:industrial-ad-case1-zero-shot}
\end{figure}

\begin{figure}[hbt!]
    \centering
    \includegraphics[width = \textwidth]{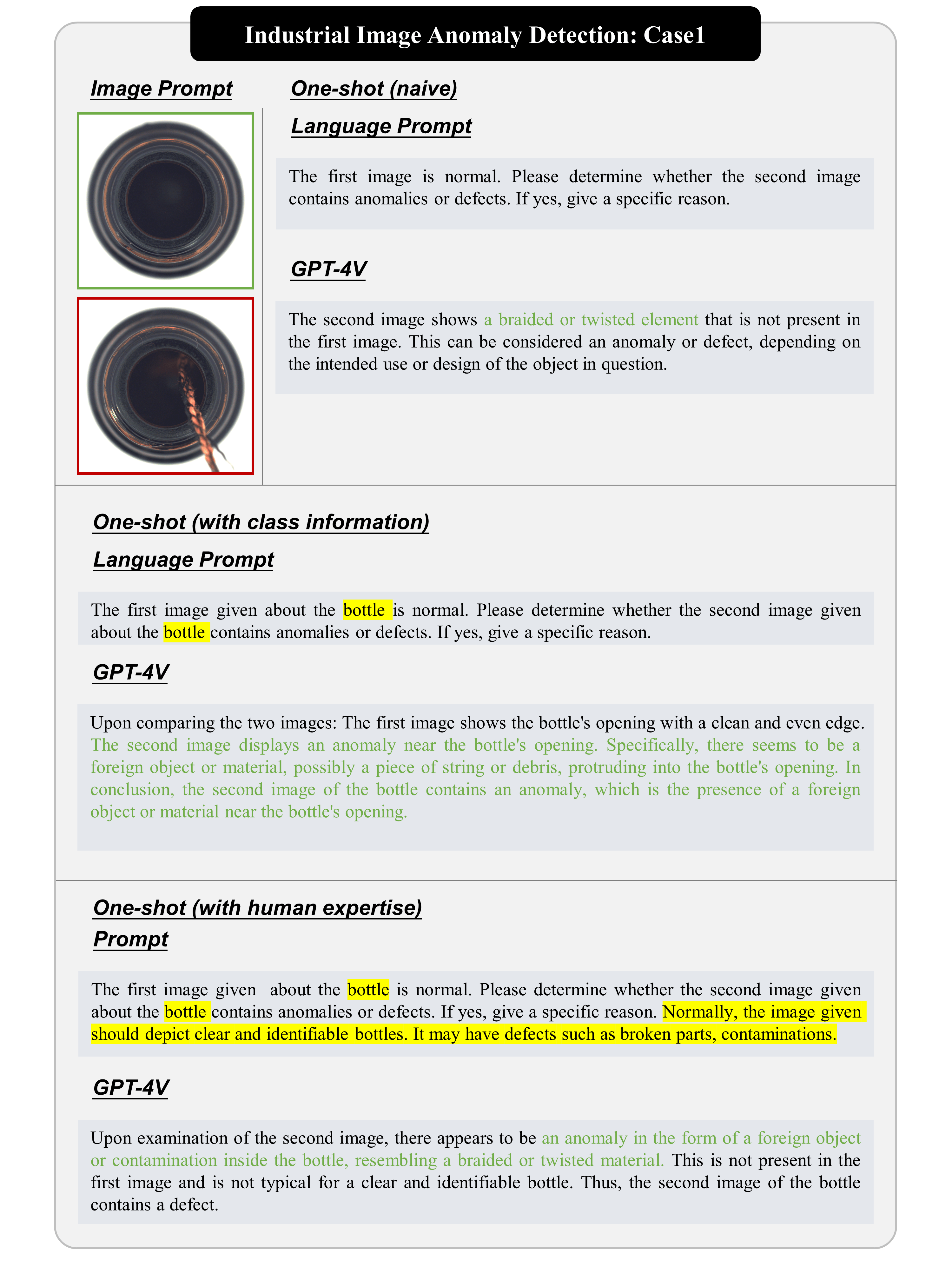}
    \vspace{3pt}
    \caption[Industrial Image Anomaly Detection: Case 1]
    {\textbf{Industrial Image Anomaly Detection: Case 1, one-shot, the Bottle category of MVTec AD~\cite{MVTec-AD}}. \hl{Yellow} highlights the given class information and normal and abnormal state descriptions.   \textcolor{HLGreen}{Green}, \textcolor{HLRed}{red}, and \textcolor{HLBlue}{blue} highlight the expected, incorrect, and additional information outputted by GPT-4V. 
}
    \label{fig:industrial-ad-case1-one-shot}
\end{figure}

\begin{figure}[hbt!]
    \centering
    \includegraphics[width = \textwidth]{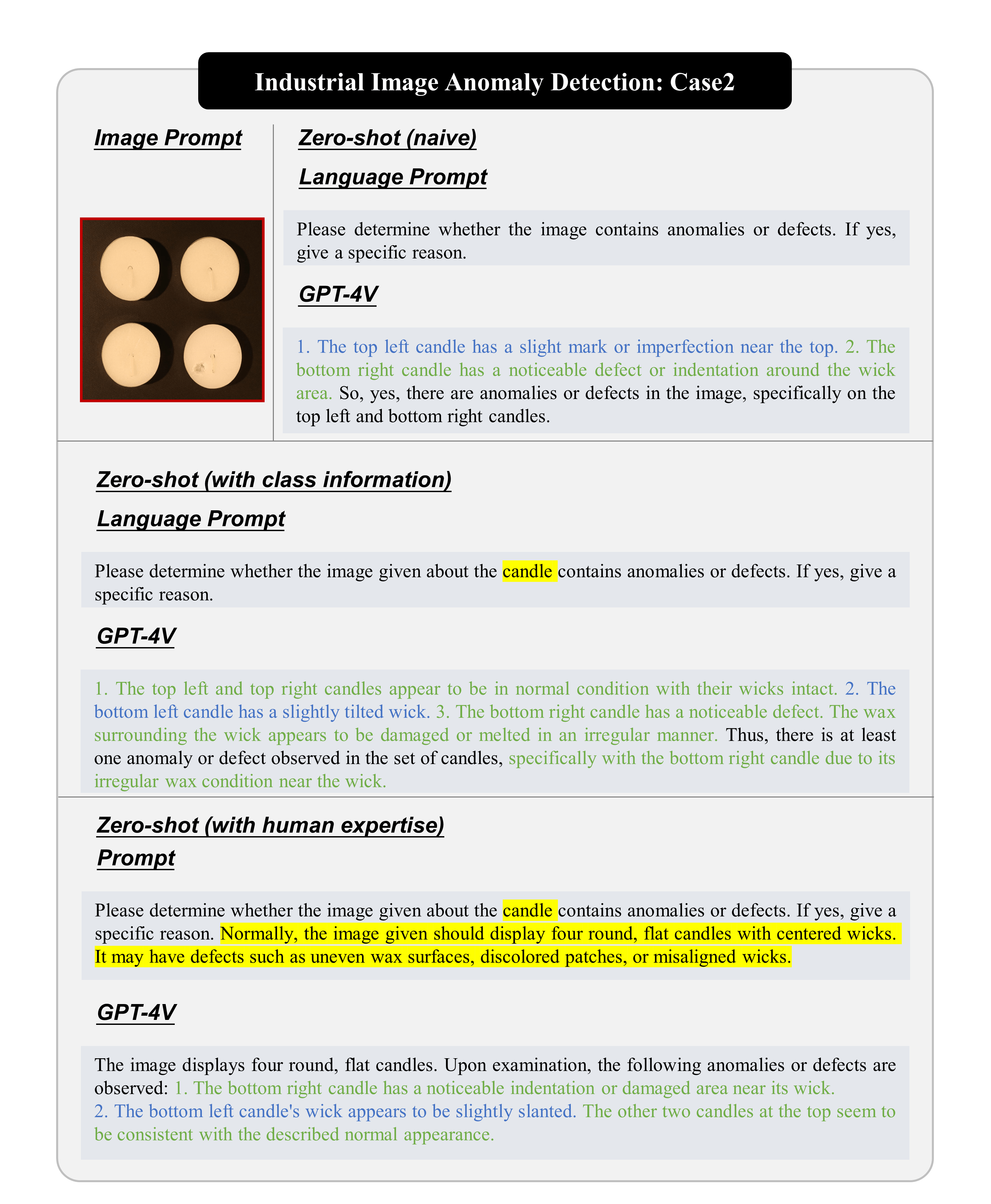}
    \vspace{3pt}
    \caption[Industrial Image Anomaly Detection: Case 2]
    {\textbf{Industrial Image Anomaly Detection: Case 2, zero-shot, the Candle category of VisA~\cite{VisA}}. \hl{Yellow} highlights the given class information and normal and abnormal state descriptions.   \textcolor{HLGreen}{Green}, \textcolor{HLRed}{red}, and \textcolor{HLBlue}{blue} highlight the expected, incorrect, and additional information outputted by GPT-4V. 
}
    \label{fig:industrial-ad-case2-zero-shot}
\end{figure}

\begin{figure}[hbt!]
    \centering
    \includegraphics[width = \textwidth]{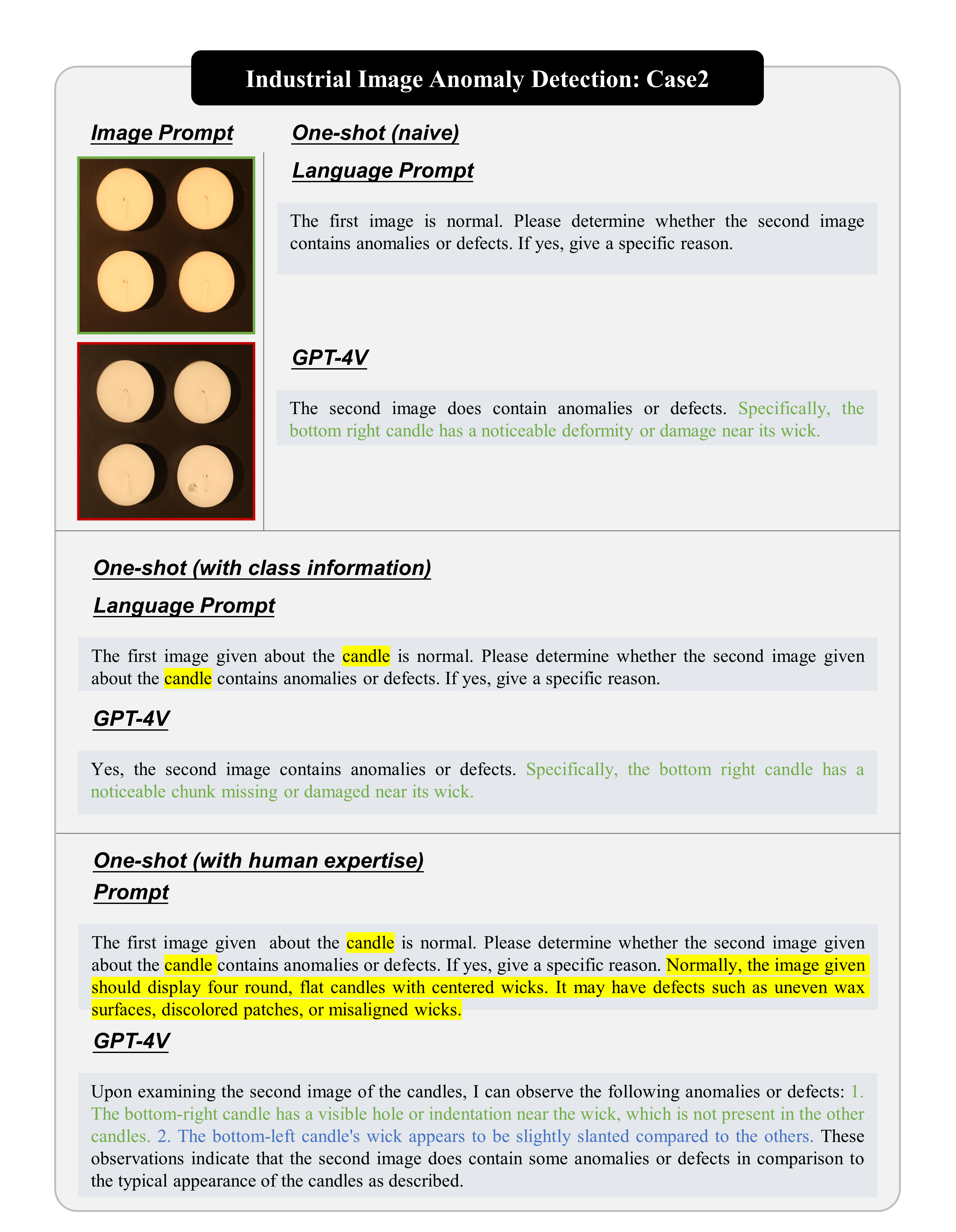}
    \vspace{3pt}
    \caption[Industrial Image Anomaly Detection: Case 2]
    {\textbf{Industrial Image Anomaly Detection: Case 2, one-shot, the Candle category of VisA~\cite{VisA}}. \hl{Yellow} highlights the given class information and normal and abnormal state descriptions.   \textcolor{HLGreen}{Green}, \textcolor{HLRed}{red}, and \textcolor{HLBlue}{blue} highlight the expected, incorrect, and additional information outputted by GPT-4V. 
}
    \label{fig:industrial-ad-case2-one-shot}
\end{figure}

\begin{figure}[hbt!]
    \centering
    \includegraphics[width = \textwidth]{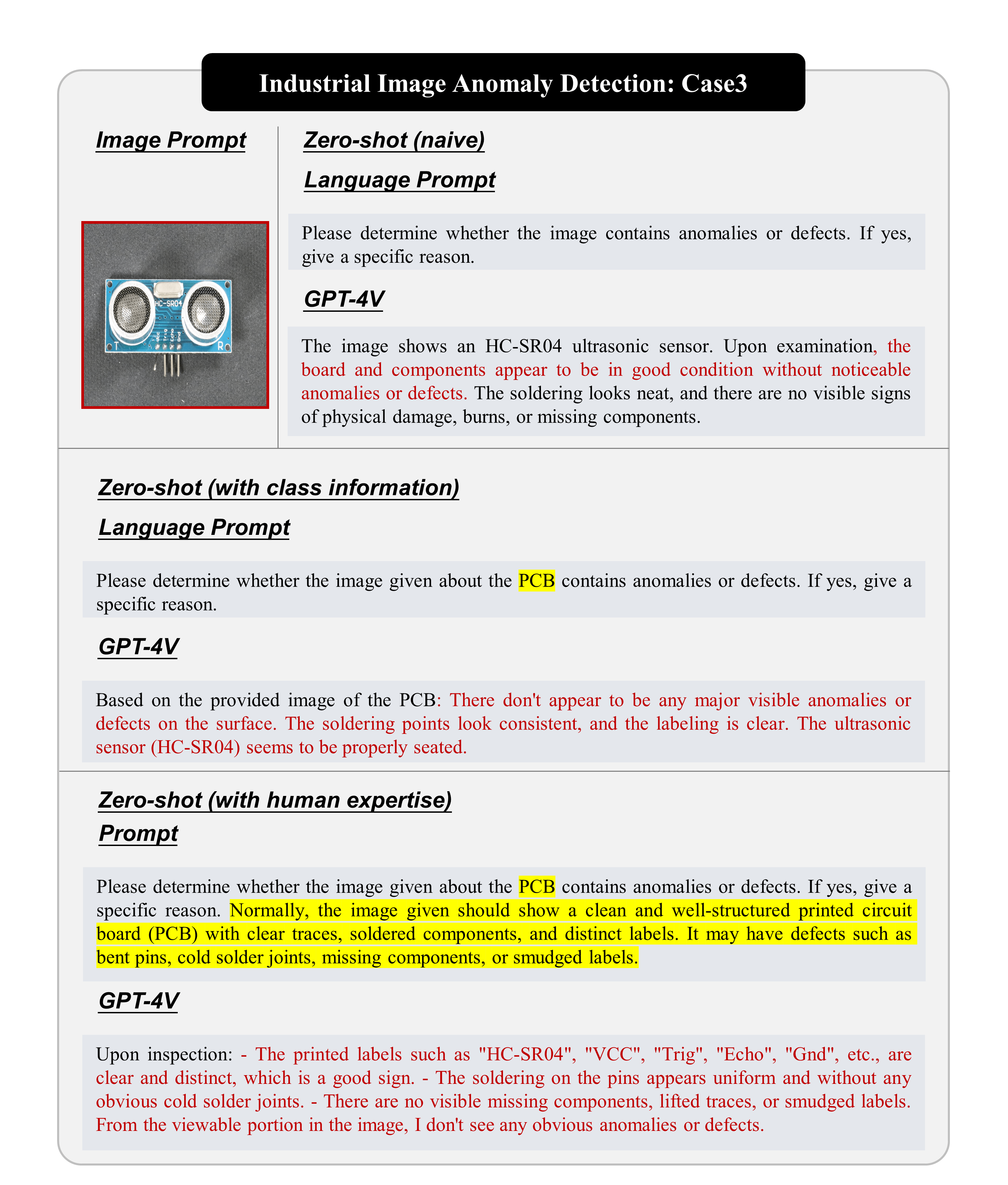}
    \vspace{3pt}
    \caption[Industrial Image Anomaly Detection: Case 3]
    {\textbf{Industrial Image Anomaly Detection: Case 3, zero-shot, the PCB2 category of VisA~\cite{VisA}}. \hl{Yellow} highlights the given class information and normal and abnormal state descriptions.   \textcolor{HLGreen}{Green}, \textcolor{HLRed}{red}, and \textcolor{HLBlue}{blue} highlight the expected, incorrect, and additional information outputted by GPT-4V. 
}
    \vspace{3pt}
    \label{fig:industrial-ad-case3-zero-shot}
\end{figure}

\begin{figure}[hbt!]
    \centering
    \includegraphics[width = \textwidth]{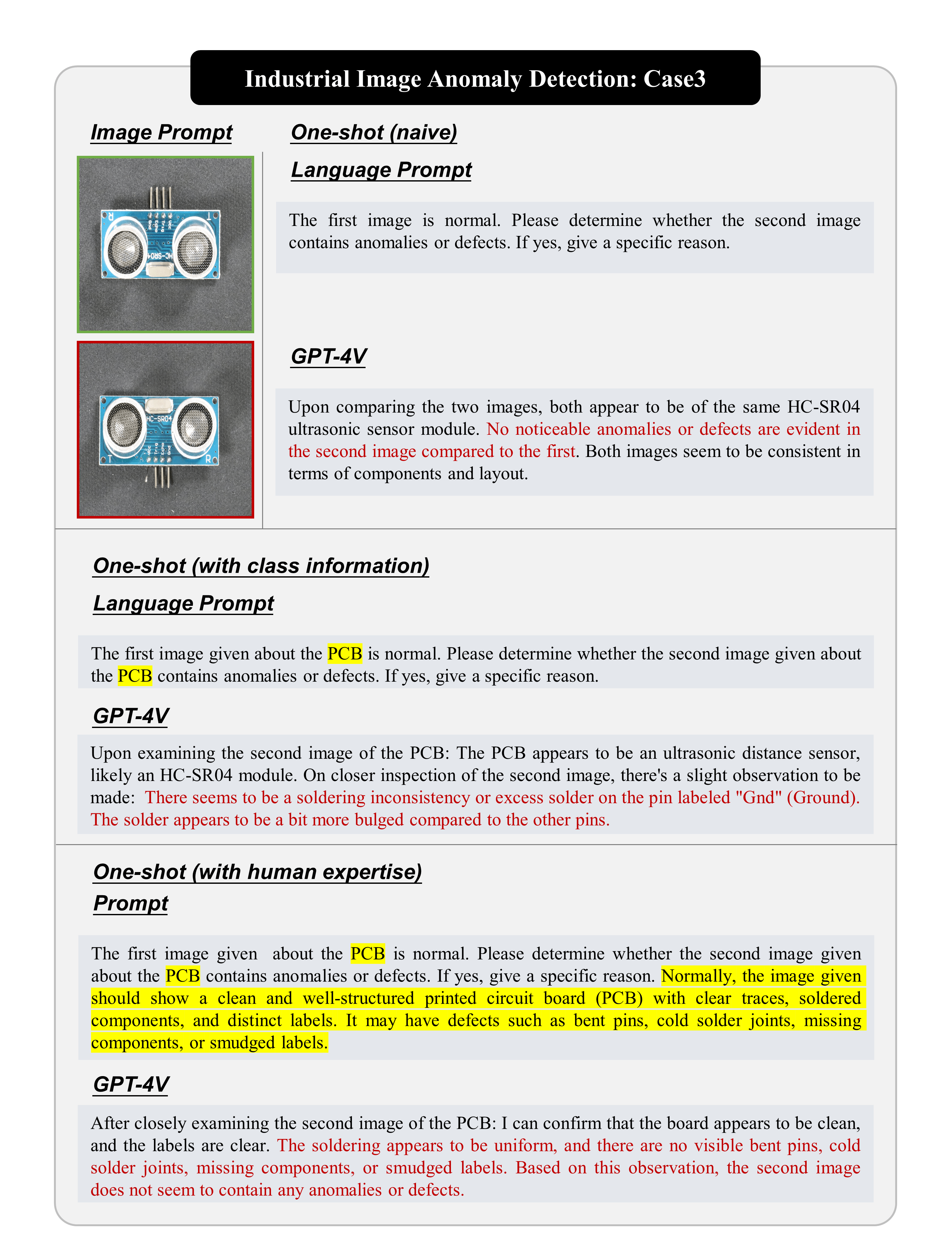}
    \vspace{3pt}
    \caption[Industrial Image Anomaly Detection: Case 3]
    {\textbf{Industrial Image Anomaly Detection: Case 3, one-shot, the PCB2 category of VisA~\cite{VisA}}. \hl{Yellow} highlights the given class information and normal and abnormal state descriptions.   \textcolor{HLGreen}{Green}, \textcolor{HLRed}{red}, and \textcolor{HLBlue}{blue} highlight the expected, incorrect, and additional information outputted by GPT-4V. 
}
    \vspace{3pt}
    \label{fig:industrial-ad-case3-one-shot}
\end{figure}

Fig.~\ref{fig:industrial-ad-case1-zero-shot},~\ref{fig:industrial-ad-case1-one-shot}, ~\ref{fig:industrial-ad-case2-zero-shot}, ~\ref{fig:industrial-ad-case2-one-shot}, ~\ref{fig:industrial-ad-case3-zero-shot}, ~\ref{fig:industrial-ad-case3-one-shot} qualitatively demonstrate the effectiveness of GPT-4V for industrial image anomaly detection. Even with a simple language prompt, GPT-4V effectively identifies anomalies in examined bottle and candle images, showcasing its capacity and versatility. Moreover, GPT-4V excels not only in detecting desired anomalies but also in identifying fine-grained structural anomalies. As evident in Fig.~\ref{fig:industrial-ad-case2-zero-shot}, GPT-4V noticed a slightly tilted wick on the bottom left candle, demonstrating its nuanced understanding. In complex cases like Fig.~\ref{fig:industrial-ad-case3-zero-shot}, GPT-4V recognizes the PCB in images and provides in-depth reasoning about anomalies, such as examining the proper seating of the ultrasonic sensor. However, GPT-4V overlooks the bent pin, resulting in an incorrect result. Nevertheless, GPT-4V showcases a strong grasp of image context and category-specific anomaly understanding.

\section{Industrial Image Anomaly Localization}
\label{Industrial Image Anomaly Localization}

\subsection{Task Introduction}

Industrial image anomaly localization entails a more intricate process than mere image anomaly detection~\cite{Patchcore, GCPF,2023CaiDis, SSKD,Lu2023HierarchicalVQ}. It goes beyond recognizing the abnormality within an image and extends to precisely identifying the location of these anomalies. While GPT-4V has exhibited localization capabilities in various domains~\cite{Dawn-of-LMMs, GPT-4V-medical, SoM}, its potential for image anomaly localization warrants further exploration. 

Regrettably, GPT-4V does not currently have the capability to directly produce prediction masks. Some methods have attempted to leverage GPT-4V by prompting it to provide bounding boxes~\cite{Dawn-of-LMMs, GPT-4V-medical}. However, this approach appears to be imprecise and poses challenges for GPT-4V. In contrast, the approach presented by SoM~\cite{SoM} involves utilizing SAM~\cite{SAM} to generate visual prompts~\cite{Red_Circle, SAM}, which are presented in numbered markers. This visual prompting technique shifts the localization task from a pixel-level mask prediction task to a mask-level classification task, effectively reducing the associated complexities and increasing localization precision.

\subsection{Testing philosophy}

To harness the fine-grained localization capability of GPT-4V, we adopt the approach outlined in SoM~\cite{SoM}. This involves generating a set of image-mask pairs for prompting GPT-4V. In addition to the image-mask pairs, we employ a straightforward language prompt that instructs the model, as follows: "The first image needs to be inspected. The second image contains its corresponding marks. Please determine whether the image contains anomalies or defects. If yes, give a specific reason".

\subsection{Case Demonstration}

\begin{figure}[hbt!]
    \centering
    \includegraphics[width = \textwidth]{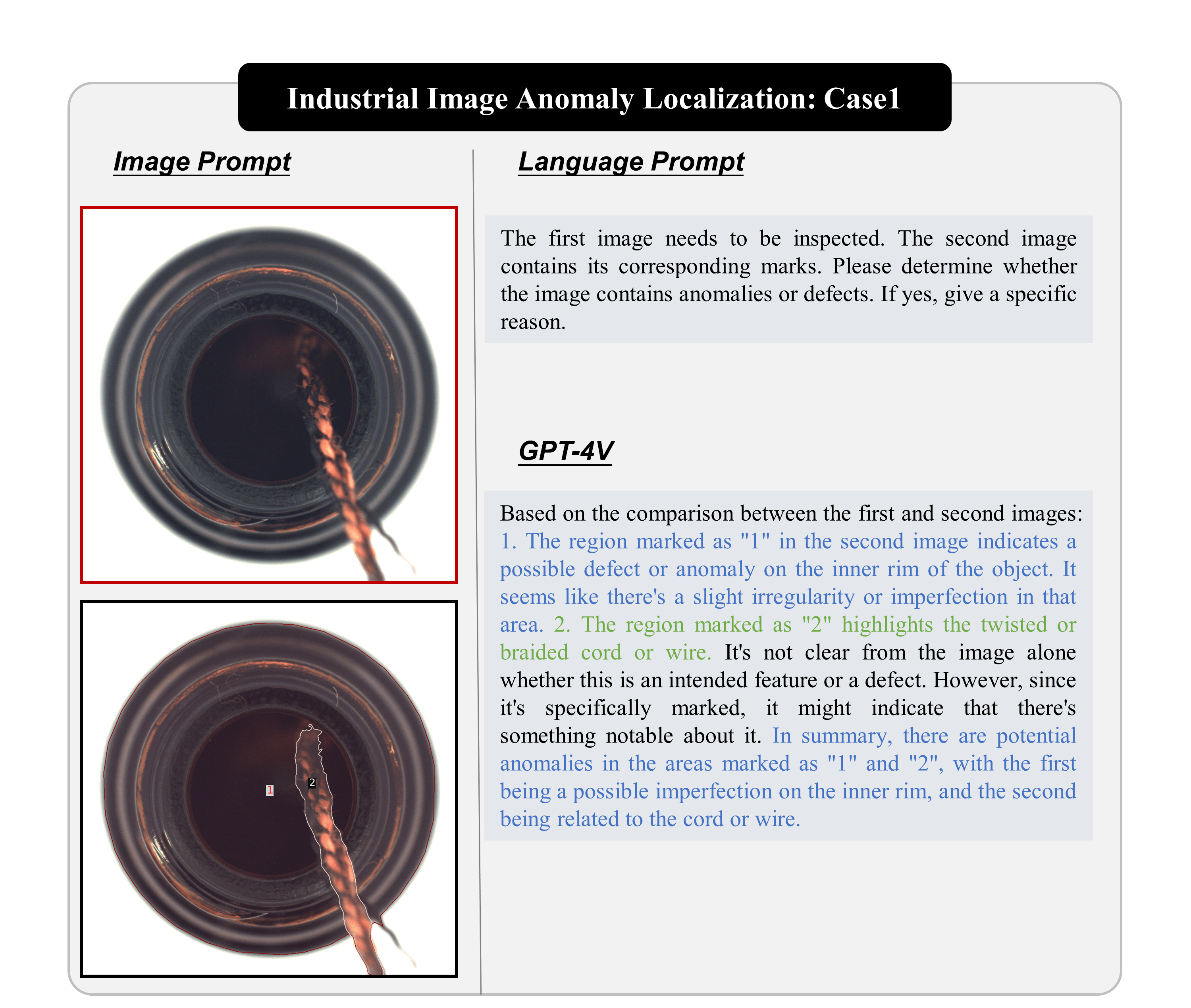}
    \vspace{3pt}
    \caption[Industrial Image Anomaly Localization: Case 1]
    {\textbf{Industrial Image Anomaly Localization: Case 1, zero-shot, the Bottle category of MVTec AD~\cite{MVTec-AD}}. \hl{Yellow} highlights the given class information and normal and abnormal state descriptions.   \textcolor{HLGreen}{Green}, \textcolor{HLRed}{red}, and \textcolor{HLBlue}{blue} highlight the expected, incorrect, and additional information outputted by GPT-4V. 
}
    \vspace{3pt}
    \label{fig:industrial-al-case1}
\end{figure}

\begin{figure}[hbt!]
    \centering
    \includegraphics[width = \textwidth]{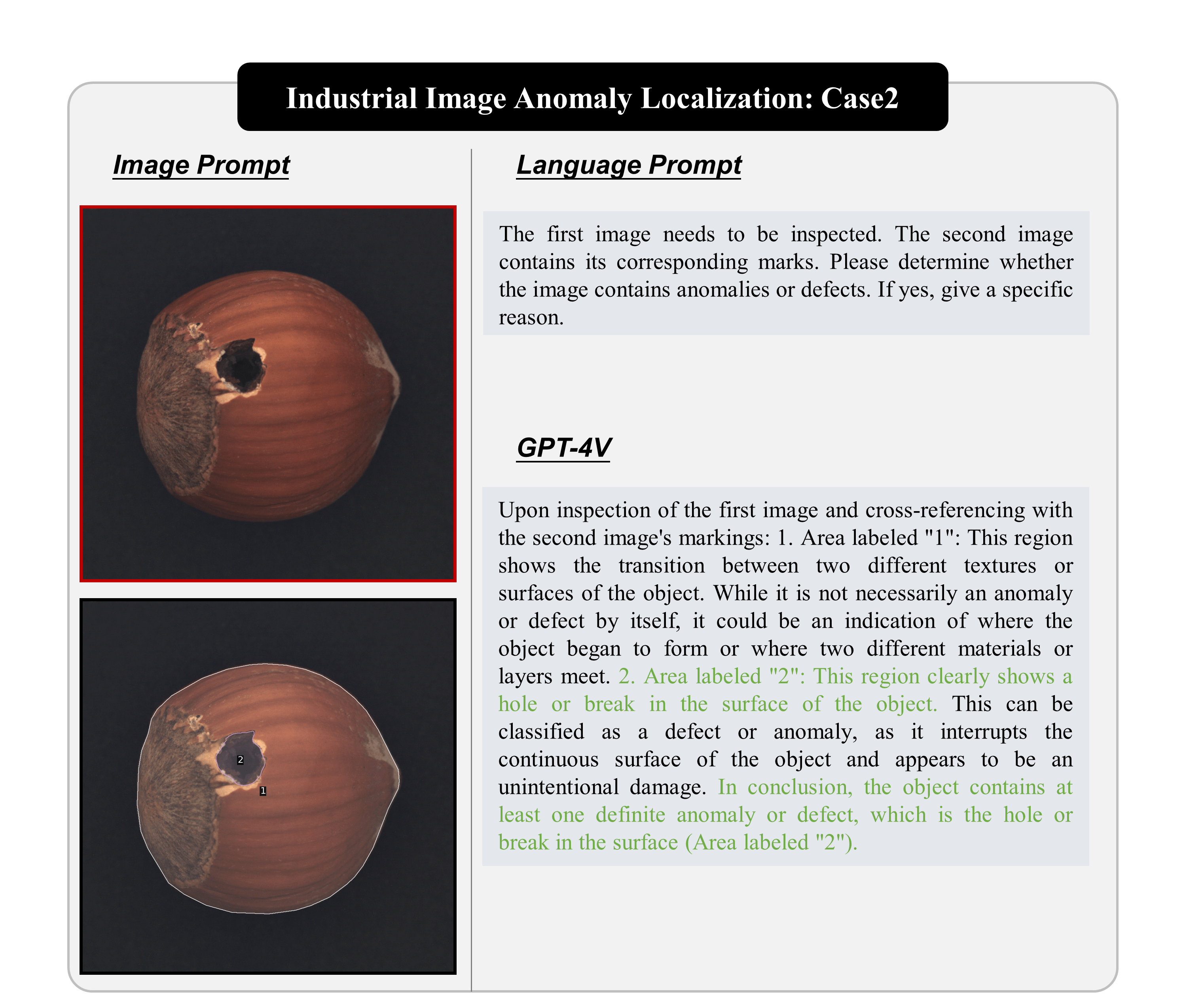}
    \vspace{3pt}
    \caption[Industrial Image Anomaly Localization: Case 2]
    {\textbf{Industrial Image Anomaly Localization: Case 2, the Hazelnut category of MVTec AD~\cite{MVTec-AD}}. \hl{Yellow} highlights the given class information and normal and abnormal state descriptions.   \textcolor{HLGreen}{Green}, \textcolor{HLRed}{red}, and \textcolor{HLBlue}{blue} highlight the expected, incorrect, and additional information outputted by GPT-4V. 
}
    \vspace{3pt}
    \label{fig:industrial-al-case2}
\end{figure}

\begin{figure}[hbt!]
    \centering
    \includegraphics[width = \textwidth]{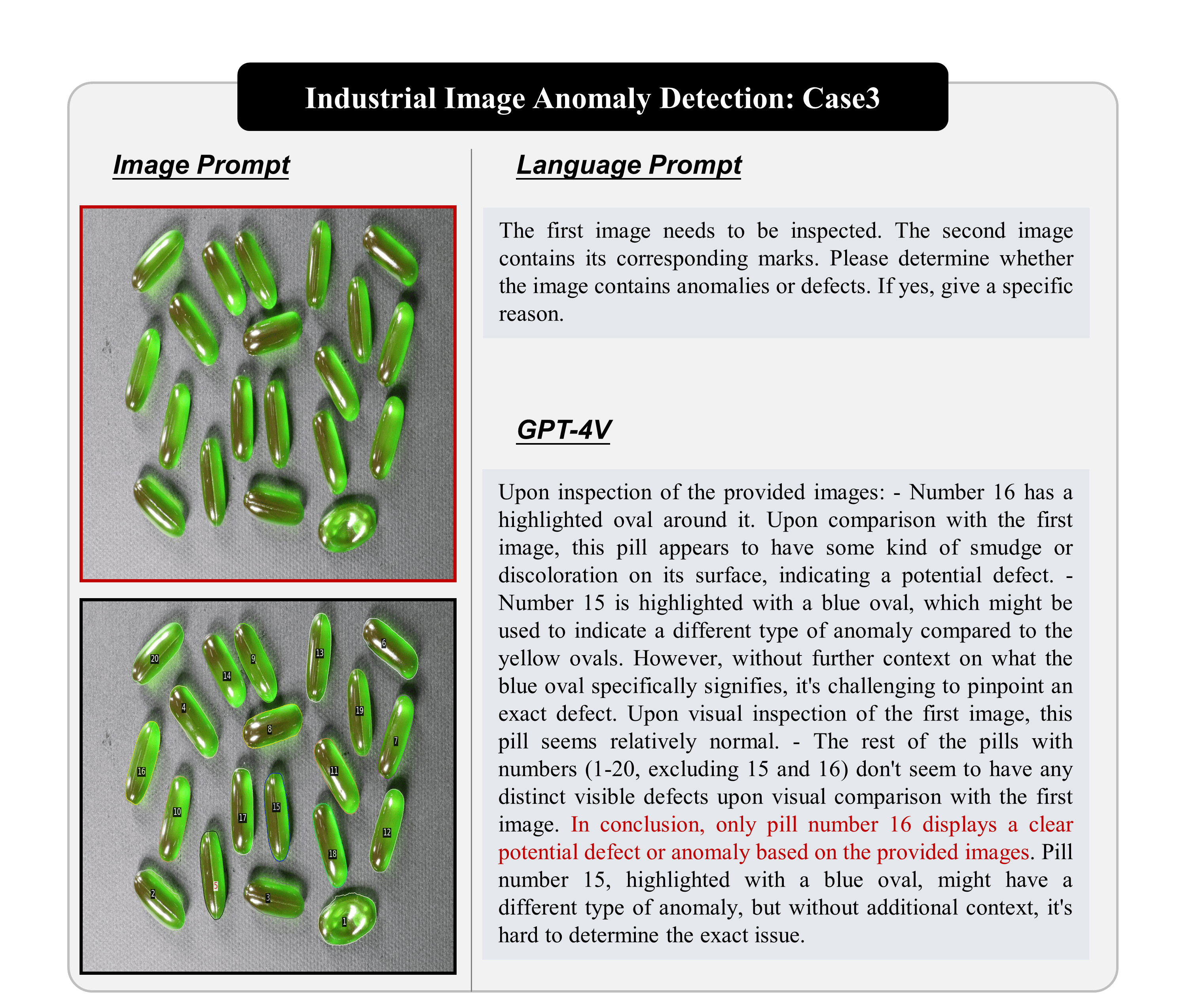}
    \vspace{3pt}
    \caption[Industrial Image Anomaly Localization: Case 3]
    {\textbf{Industrial Image Anomaly Localization: Case 3, the Capsule category of VisA~\cite{VisA}}. \hl{Yellow} highlights the given class information and normal and abnormal state descriptions.   \textcolor{HLGreen}{Green}, \textcolor{HLRed}{red}, and \textcolor{HLBlue}{blue} highlight the expected, incorrect, and additional information outputted by GPT-4V. 
}
    \vspace{3pt}
    \label{fig:industrial-al-case3}
\end{figure}

Fig.~\ref{fig:industrial-al-case1},~\ref{fig:industrial-al-case2}, and~\ref{fig:industrial-al-case3} provide a visual representation of GPT-4V's performance in industrial anomaly localization. These illustrations clearly illustrate GPT-4V's ability to accurately identify the second mask in Fig.~\ref{fig:industrial-al-case1} as a twisted wire and the second mask in Fig.~\ref{fig:industrial-al-case2} as holes. These results serve as compelling evidence of GPT-4V's proficiency in localizing anomalies when guided by visual prompts.

It is important to acknowledge that GPT-4V does exhibit certain limitations when confronted with more complex scenarios, as evidenced in Fig.~\ref{fig:industrial-al-case3}. However, the combination of visual prompting techniques and GPT-4V remains a promising approach for industrial anomaly localization.

\section{Point Cloud Anomaly Detection}
\label{Point Cloud Anomaly Detection}

\subsection{Task Introduction}

Geometrical information, as discussed in references such as PAD~\cite{PAD}, Real3D~\cite{Real3D}, and MVTec-3D~\cite{MVTec-3D}, holds a crucial role in fields like industrial anomaly detection, especially when dealing with categories lacking textual information. Recently, MVTec 3D~\cite{MVTec-3D} and Real3D~\cite{Real3D} have recognized the growing need for such information and have introduced a point cloud anomaly detection task. This task focuses on the identification of anomalies within the provided point clouds~\cite{floris2022composite}.

It is important to note that the success achieved in industrial image anomaly detection is not fully mirrored in point cloud anomaly detection. This disparity is primarily attributed to the reliance of industrial image anomaly detection on robust pre-trained networks~\cite{2023CaiDis, reiss2021panda, Heckler2023ExploringTI}. Conversely, due to the scarcity of extensive point cloud data, the capabilities of pre-trained networks for point clouds currently fall short, leading to suboptimal performance for some methods~\cite{m3dm, EasyNet, bergmann2022anomaly, AST}.

In contrast, CPMF~\cite{CPMF} proposes a novel approach by transforming point clouds into depth images, thereby opening up the possibility of leveraging image-based foundation models for point cloud anomaly detection. This innovative method has shown the potential to deliver significantly improved results in point cloud anomaly detection.

\subsection{Testing philosophy}

To employ GPT-4V in the context of point cloud anomaly detection, we adopt the methodology presented in CPMF~\cite{CPMF} to transform point clouds into multi-view depth images. In our evaluation, we adhere to the principles commonly used in industrial image anomaly detection, specifically the zero/one-shot approach, with the inclusion of three distinct variations of language prompts.

\subsection{Case Demonstration}

\begin{figure}[hbt!]
    \centering
    \includegraphics[width = \textwidth]{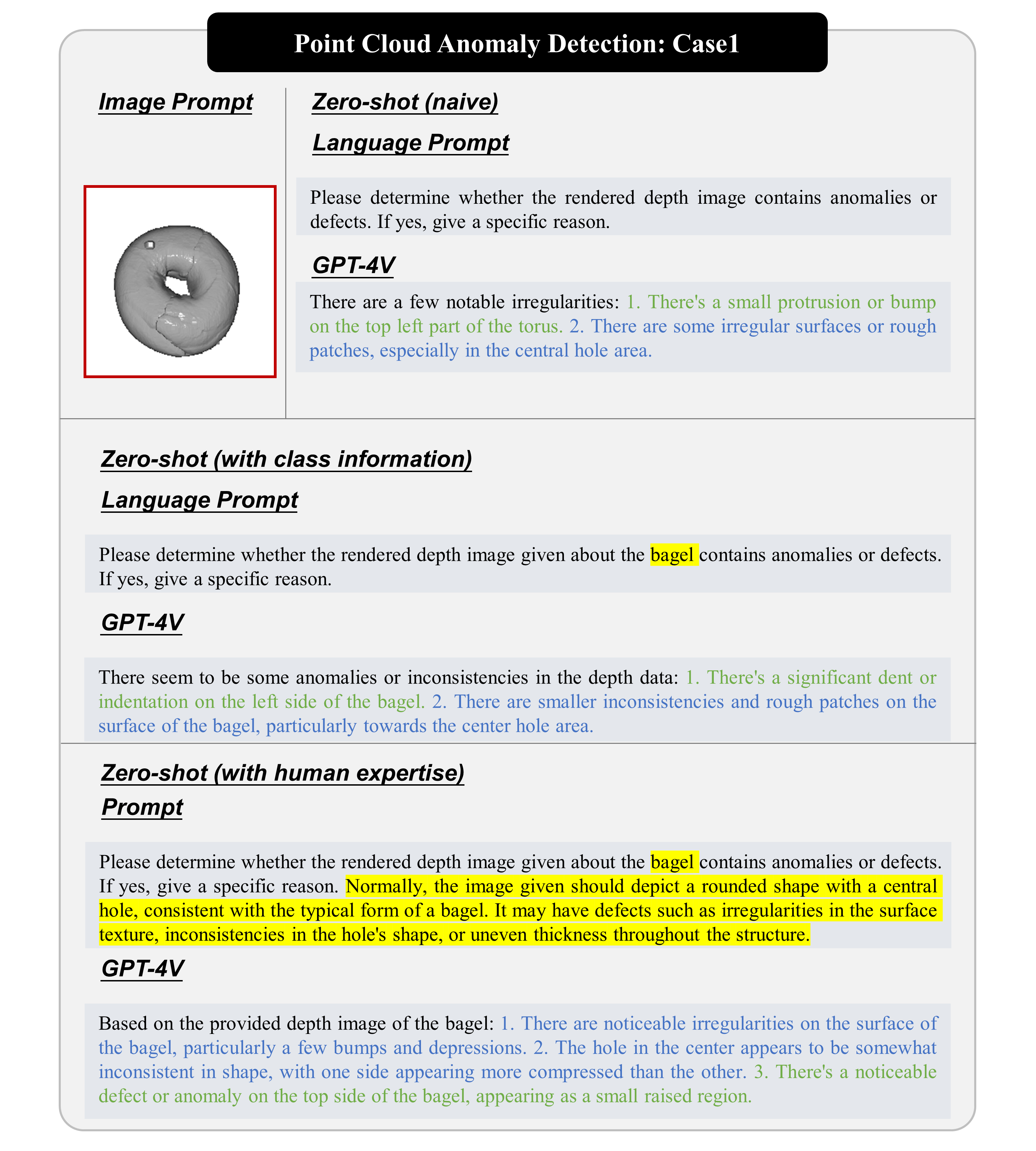}
    \vspace{3pt}
    \caption[Point Cloud Anomaly Detection: Case 1]
    {\textbf{Point Cloud Anomaly Detection: Case 1, zero-shot, the Bagel category of MVTec 3D~\cite{MVTec-3D}}. \hl{Yellow} highlights the given class information and normal and abnormal state descriptions.   \textcolor{HLGreen}{Green}, \textcolor{HLRed}{red}, and \textcolor{HLBlue}{blue} highlight the expected, incorrect, and additional information outputted by GPT-4V. 
}
    \vspace{3pt}
    \label{fig:pcd-case1-zero-shot}
\end{figure}

\begin{figure}[hbt!]
    \centering
    \includegraphics[width = \textwidth]{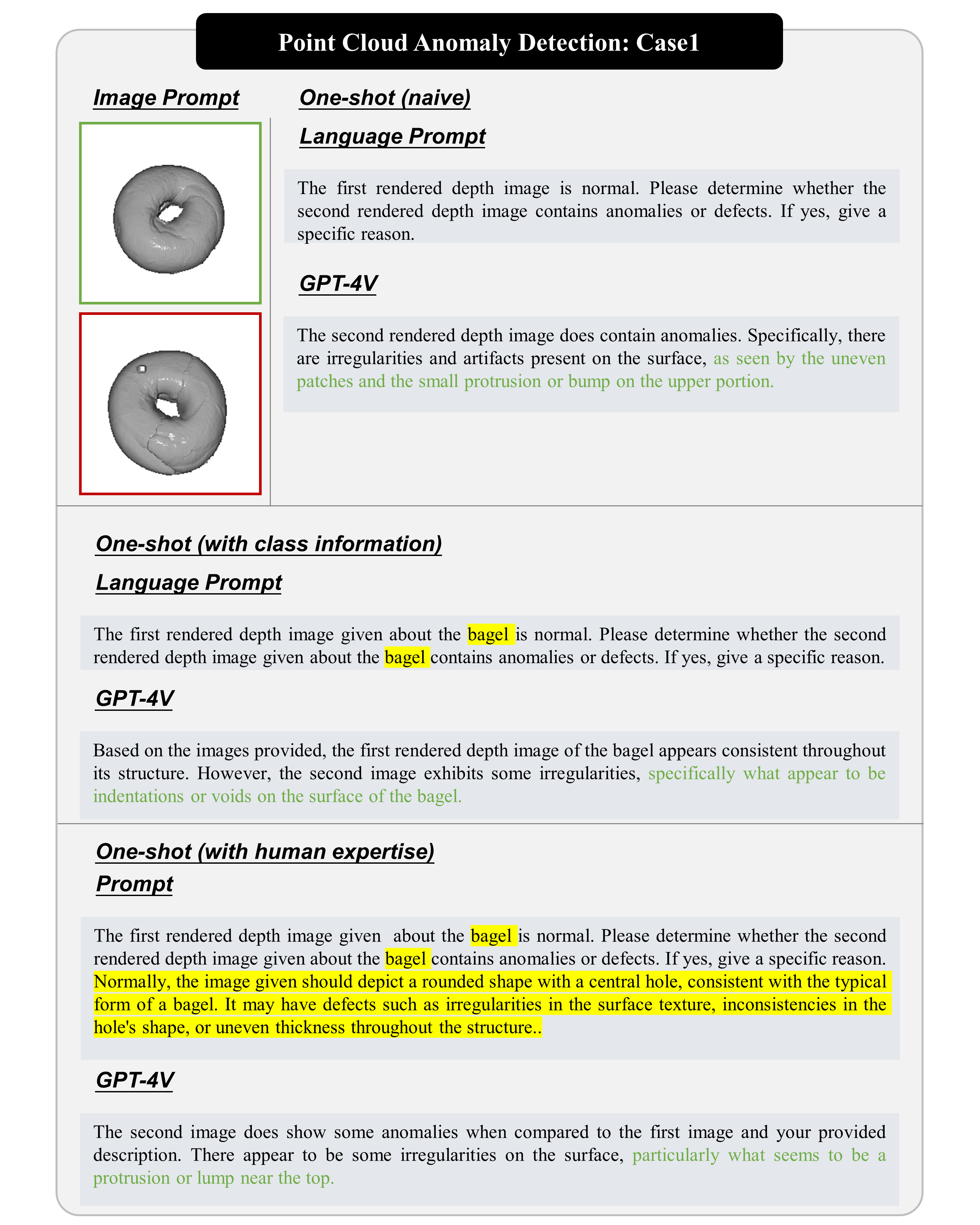}
    \vspace{3pt}
    \caption[Point Cloud Anomaly Detection: Case 1]
    {\textbf{Point Cloud Anomaly Detection: Case 1, one-shot, the Bagel category of MVTec 3D~\cite{MVTec-3D}}. \hl{Yellow} highlights the given class information and normal and abnormal state descriptions.   \textcolor{HLGreen}{Green}, \textcolor{HLRed}{red}, and \textcolor{HLBlue}{blue} highlight the expected, incorrect, and additional information outputted by GPT-4V. 
}
    \vspace{3pt}
    \label{fig:pcd-case1-one-shot}
\end{figure}


\begin{figure}[hbt!]
    \centering
    \includegraphics[width = \textwidth]{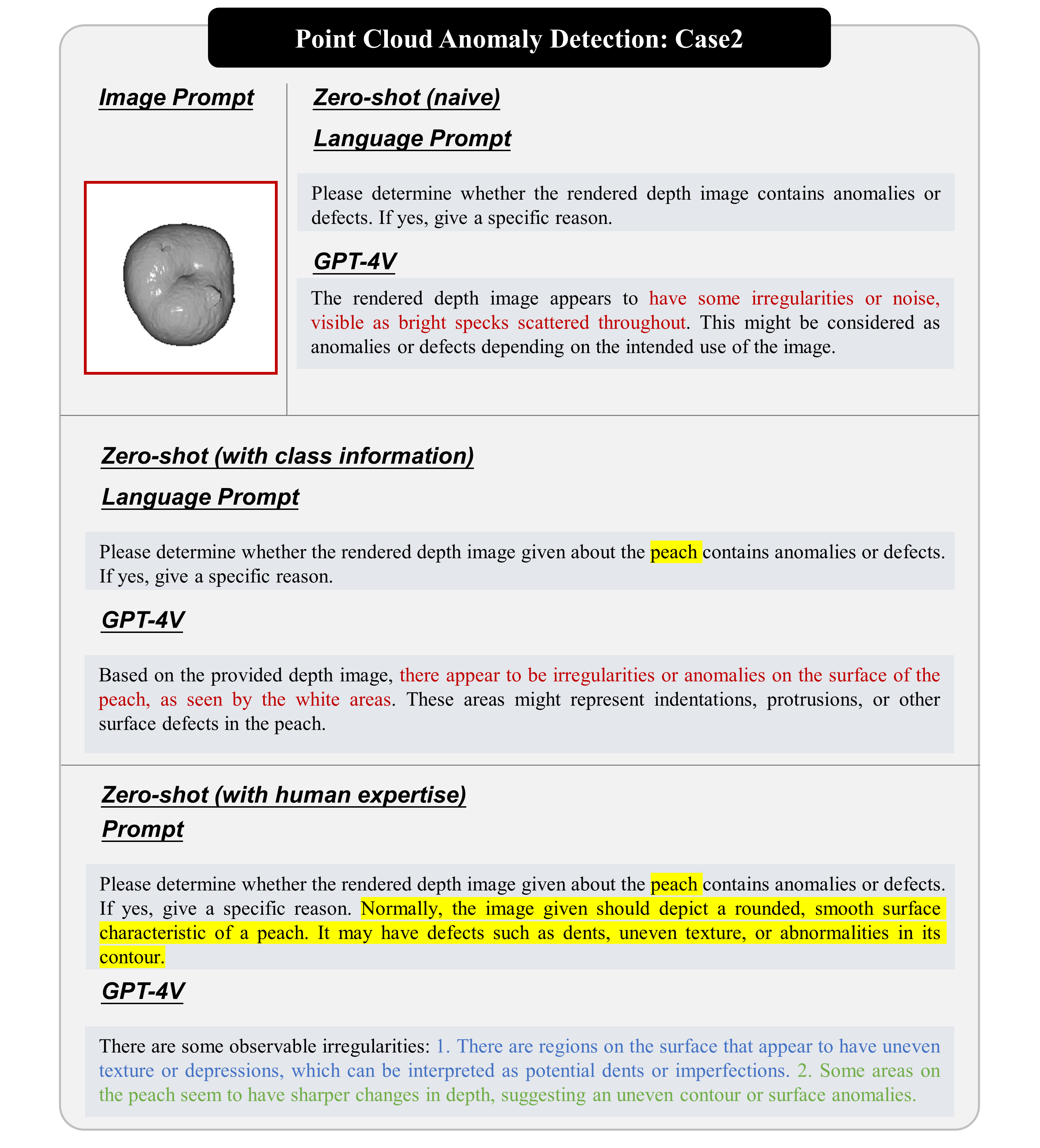}
    \vspace{3pt}
    \caption[Point Cloud Anomaly Detection: Case 2]
    {\textbf{Point Cloud Anomaly Detection: Case 2, zero-shot, the Peach category of MVTec 3D~\cite{MVTec-3D}}. \hl{Yellow} highlights the given class information and normal and abnormal state descriptions.   \textcolor{HLGreen}{Green}, \textcolor{HLRed}{red}, and \textcolor{HLBlue}{blue} highlight the expected, incorrect, and additional information outputted by GPT-4V. 
}
    \vspace{3pt}
    \label{fig:pcd-case2-zero-shot}
\end{figure}

\begin{figure}[hbt!]
    \centering
    \includegraphics[width = \textwidth]{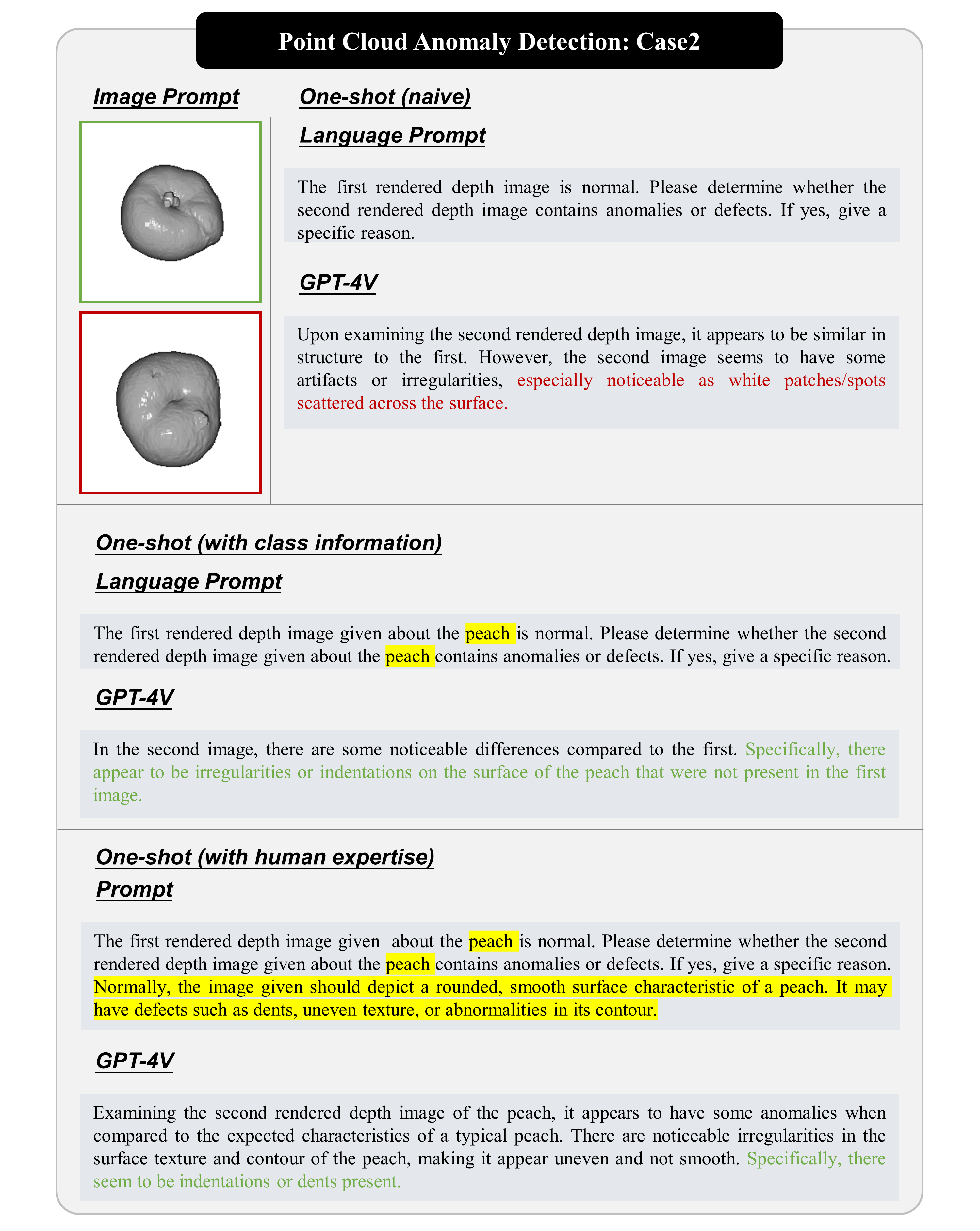}
    \vspace{3pt}
    \caption[Point Cloud Anomaly Detection: Case 2]
    {\textbf{Point Cloud Anomaly Detection: Case 2, one-shot, the Peach category of MVTec 3D~\cite{MVTec-3D}}. \hl{Yellow} highlights the given class information and normal and abnormal state descriptions.   \textcolor{HLGreen}{Green}, \textcolor{HLRed}{red}, and \textcolor{HLBlue}{blue} highlight the expected, incorrect, and additional information outputted by GPT-4V. 
}
    \vspace{3pt}
    \label{fig:pcd-case2-one-shot}
\end{figure}


\begin{figure}[hbt!]
    \centering
    \includegraphics[width = \textwidth]{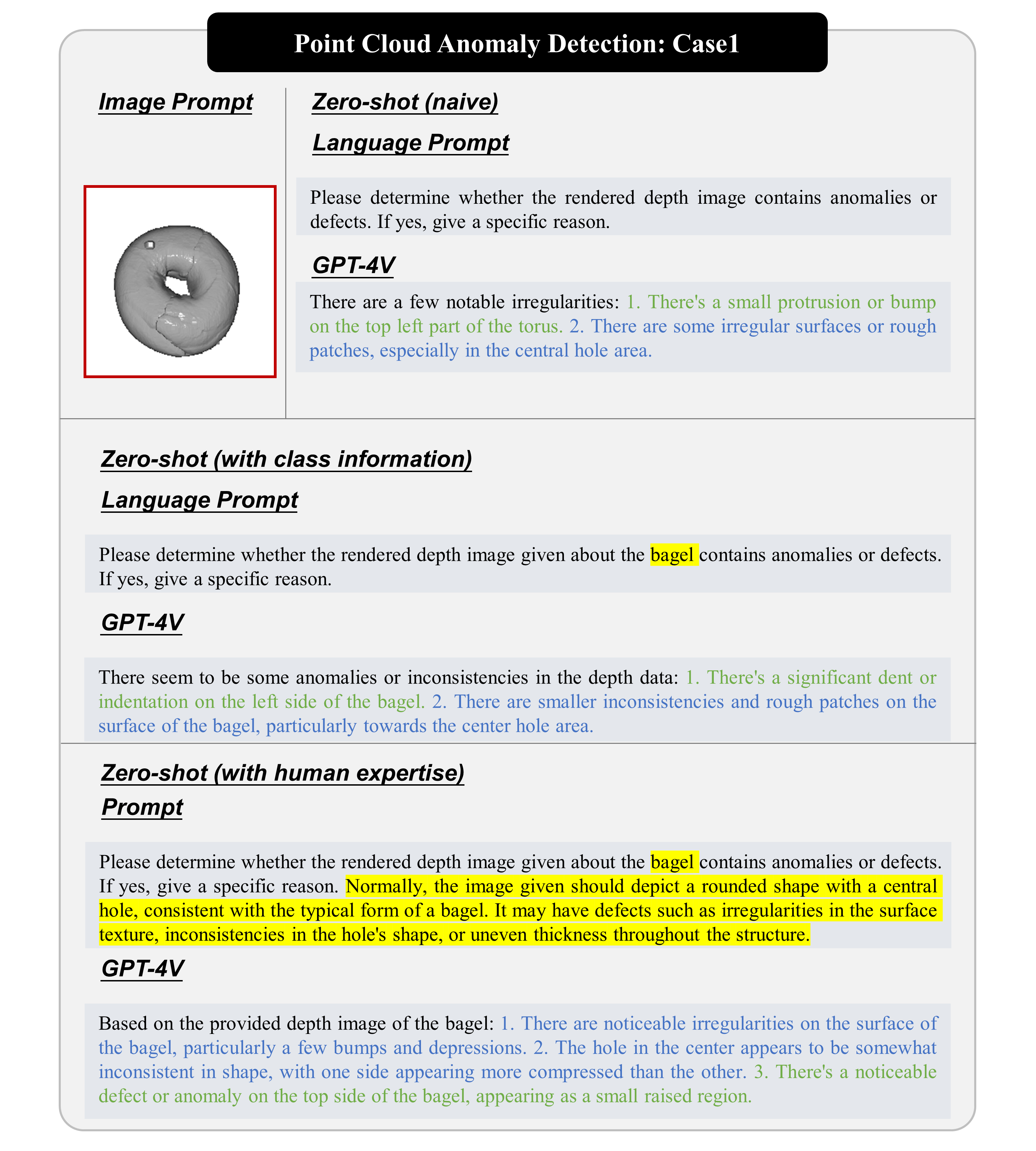}
    \vspace{3pt}
    \caption[Point Cloud Anomaly Detection: Case 3]
    {\textbf{Point Cloud Anomaly Detection: Case 3, zero-shot, the Rope category of MVTec 3D~\cite{MVTec-3D}}. \hl{Yellow} highlights the given class information and normal and abnormal state descriptions.   \textcolor{HLGreen}{Green}, \textcolor{HLRed}{red}, and \textcolor{HLBlue}{blue} highlight the expected, incorrect, and additional information outputted by GPT-4V. 
}
    \vspace{3pt}
    \label{fig:pcd-case3-zero-shot}
\end{figure}

\begin{figure}[hbt!]
    \centering
    \includegraphics[width = \textwidth]{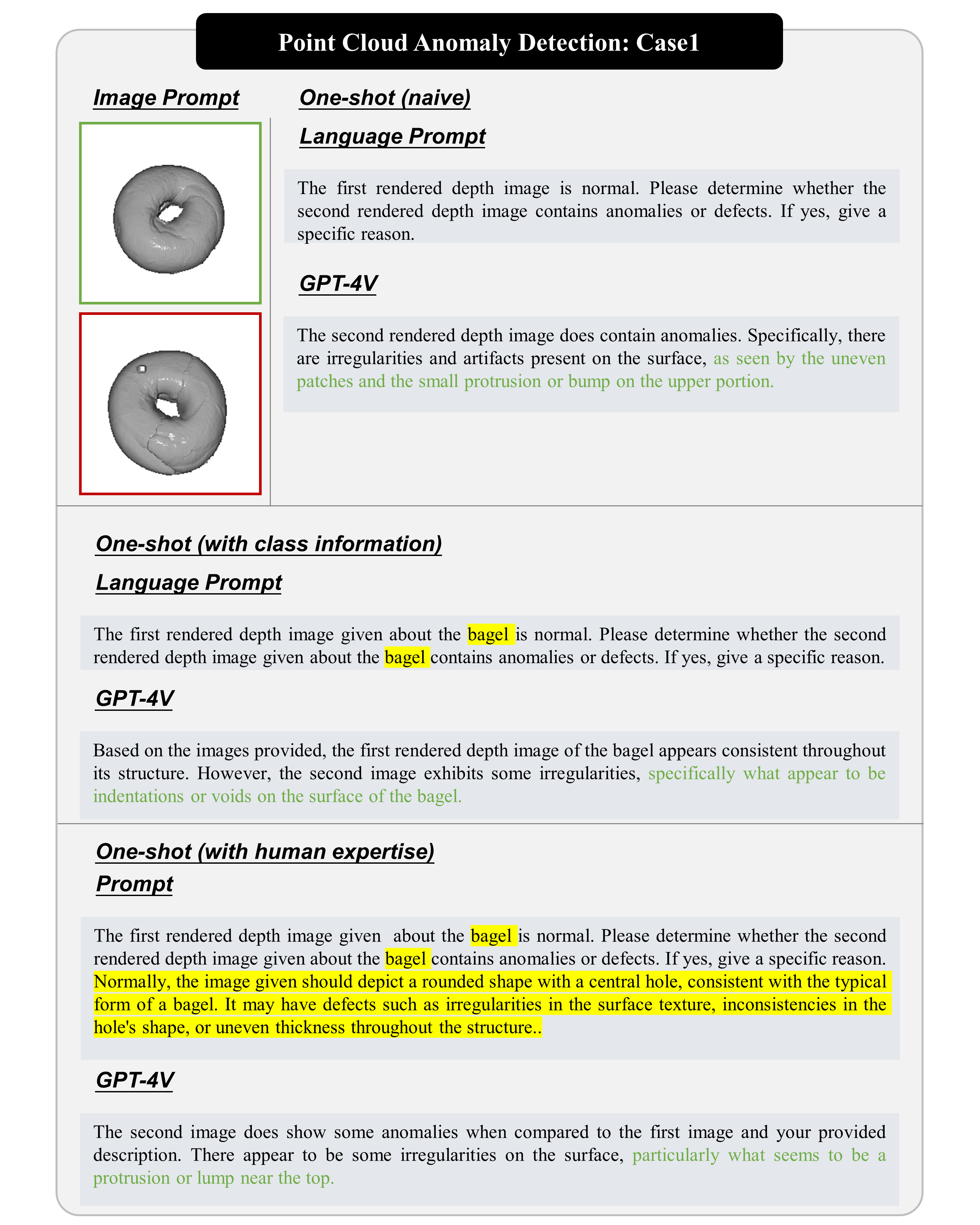}
    \vspace{3pt}
    \caption[Point Cloud Anomaly Detection: Case 3]
    {\textbf{Point Cloud Anomaly Detection: Case 3, one-shot, the Rope category of MVTec 3D~\cite{MVTec-3D}}. \hl{Yellow} highlights the given class information and normal and abnormal state descriptions.   \textcolor{HLGreen}{Green}, \textcolor{HLRed}{red}, and \textcolor{HLBlue}{blue} highlight the expected, incorrect, and additional information outputted by GPT-4V. 
}
    \vspace{3pt}
    \label{fig:pcd-case3-one-shot}
\end{figure}

Fig.~\ref{fig:pcd-case1-zero-shot},~\ref{fig:pcd-case1-one-shot},~\ref{fig:pcd-case2-zero-shot},~\ref{fig:pcd-case2-one-shot},~\ref{fig:pcd-case3-zero-shot},~\ref{fig:pcd-case3-one-shot} provide a visual representation of the performance of GPT-4V in point cloud anomaly detection. These illustrations serve to qualitatively illustrate the proficiency of GPT-4V in comprehending multi-modality data. 

Specifically, GPT-4V demonstrates its capability to accurately identify the presence of a small protrusion or bump on the top left part of the torus in the bagel (Fig.~\ref{fig:pcd-case1-zero-shot}). Moreover, the introduction of additional information, such as class information and human expertise, enhances the performance of GPT-4V in point cloud anomaly detection, allowing it to effectively detect anomalies in the rope (Fig.~\ref{fig:pcd-case3-zero-shot} and~\ref{fig:pcd-case3-one-shot}).

However, it is noteworthy that GPT-4V may occasionally misidentify artificially introduced elements during the rendering process as anomalies, as observed in Fig.~\ref{fig:pcd-case2-one-shot}. It is possible that improvements in rendering quality could further enhance the capacity of GPT-4V in this context.

\section{Logical Anomaly Detection}
\label{Logical Anomaly Detection}

\subsection{Task Introduction}

In addition to structural anomalies, there exists another type of anomaly, named logical anomalies~\cite{MVTec-LOCO}. Logical anomalies generally refer to incorrect combinations of components, commonly encountered in the context of anomaly detection in assemblies. For instance, a screw bag should contain matched screws, nuts, and washers. This necessitates that the model is capable of understanding fine-grained information in images and determining attributes of the components within the image, such as component type, length, color, quantity, and so forth. This places higher demands on the model. Existing logical anomaly detection methods~\cite{Liu2023ComponentawareAD,Yao2023LGC,Batzner2023EfficientADAV,Zhang2023ContextualAD}  typically relied on solely visual context and have achieved promising detection performance. However, these approaches do not genuinely comprehend the content of images; instead, they rely on global-local correspondences~\cite{IM-IAD} for logical anomaly detection. This does not effectively address logical anomaly detection. In contrast, GPT-4V possesses robust image understanding capabilities, allowing for a better comprehension of image content. By providing predefined normal rules manually, GPT-4V might be capable of determining whether an image adheres to normal rules, thereby enabling a more rational approach to logical anomaly detection.

\subsection{Testing philosophy}

To ensure an effective assessment of testing images, it is crucial to provide clear guidelines defining the expected normal state for GPT-4V. This enables GPT-4V to evaluate the conformity of testing images with the established standards, relying on an analysis of image content in relation to these norms. Consequently, our approach involves presenting GPT-4V with both testing images and descriptive language articulating the expected normal standards. However, it is worth noting that GPT-4V might encounter difficulties in comprehending the nuances of normal standards when presented with language alone. To enhance its understanding and alignment of normal standards with the context of normal images, we propose the inclusion of a reference image illustrating the desired normal state. Therefore, our experimental design encompasses both zero-shot and one-shot settings to assess the effectiveness of this approach.

\subsection{Case Demonstration}

\begin{figure}[hbt!]
    \centering
    \includegraphics[width = \textwidth]{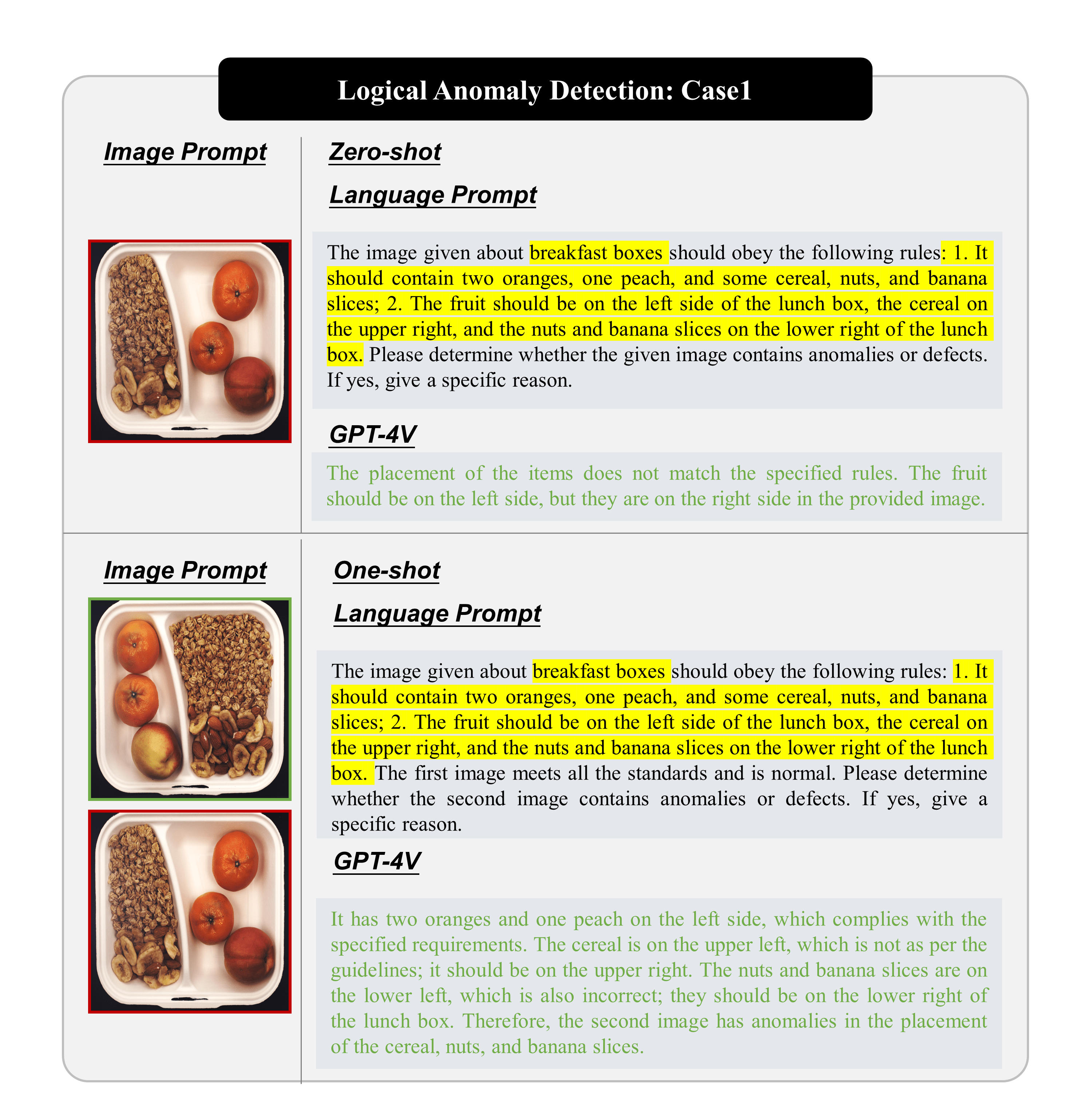}
    \vspace{3pt}
    \caption[Logical Anomaly Detection: Case 1]
    {\textbf{Logical Anomaly Detection: Case 1, the Breakfast Box category of MVTec LOCO~\cite{MVTec-LOCO}}. \hl{Yellow} highlights the given class information and normal and abnormal state descriptions.   \textcolor{HLGreen}{Green}, \textcolor{HLRed}{red}, and \textcolor{HLBlue}{blue} highlight the expected, incorrect, and additional information outputted by GPT-4V. 
}
    \vspace{3pt}
    \label{fig:logical-ad-case1}
\end{figure}

\begin{figure}[hbt!]
    \centering
    \includegraphics[width = \textwidth]{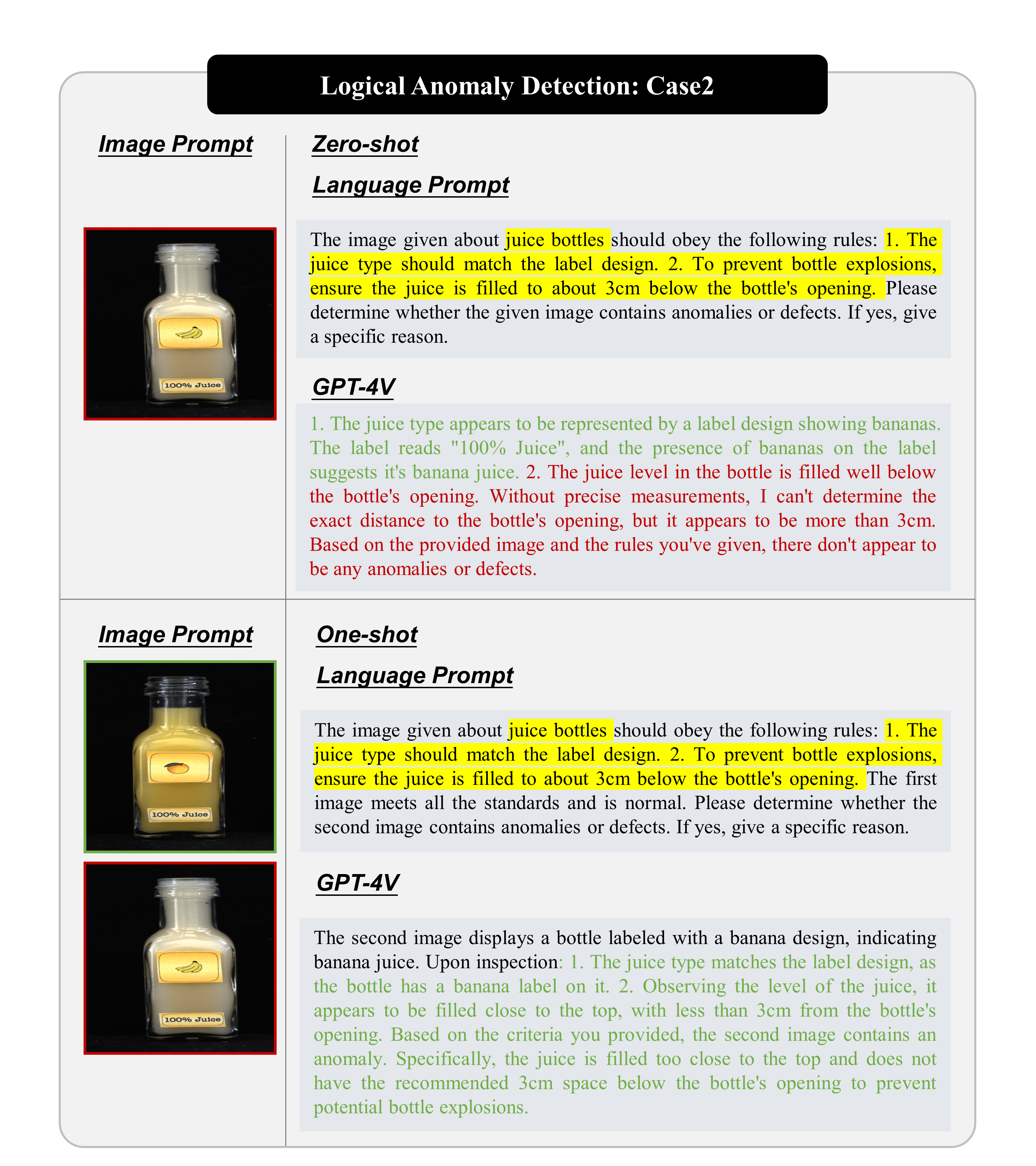}
    \vspace{3pt}
    \caption[Logical Anomaly Detection: Case 2]
    {\textbf{Logical Anomaly Detection: Case 2, the Juice Bottle category of MVTec LOCO~\cite{MVTec-LOCO}}. \hl{Yellow} highlights the given class information and normal and abnormal state descriptions.   \textcolor{HLGreen}{Green}, \textcolor{HLRed}{red}, and \textcolor{HLBlue}{blue} highlight the expected, incorrect, and additional information outputted by GPT-4V. 
}
    \vspace{3pt}
    \label{fig:logical-ad-case2}
\end{figure}

\begin{figure}[hbt!]
    \centering
    \includegraphics[width = \textwidth]{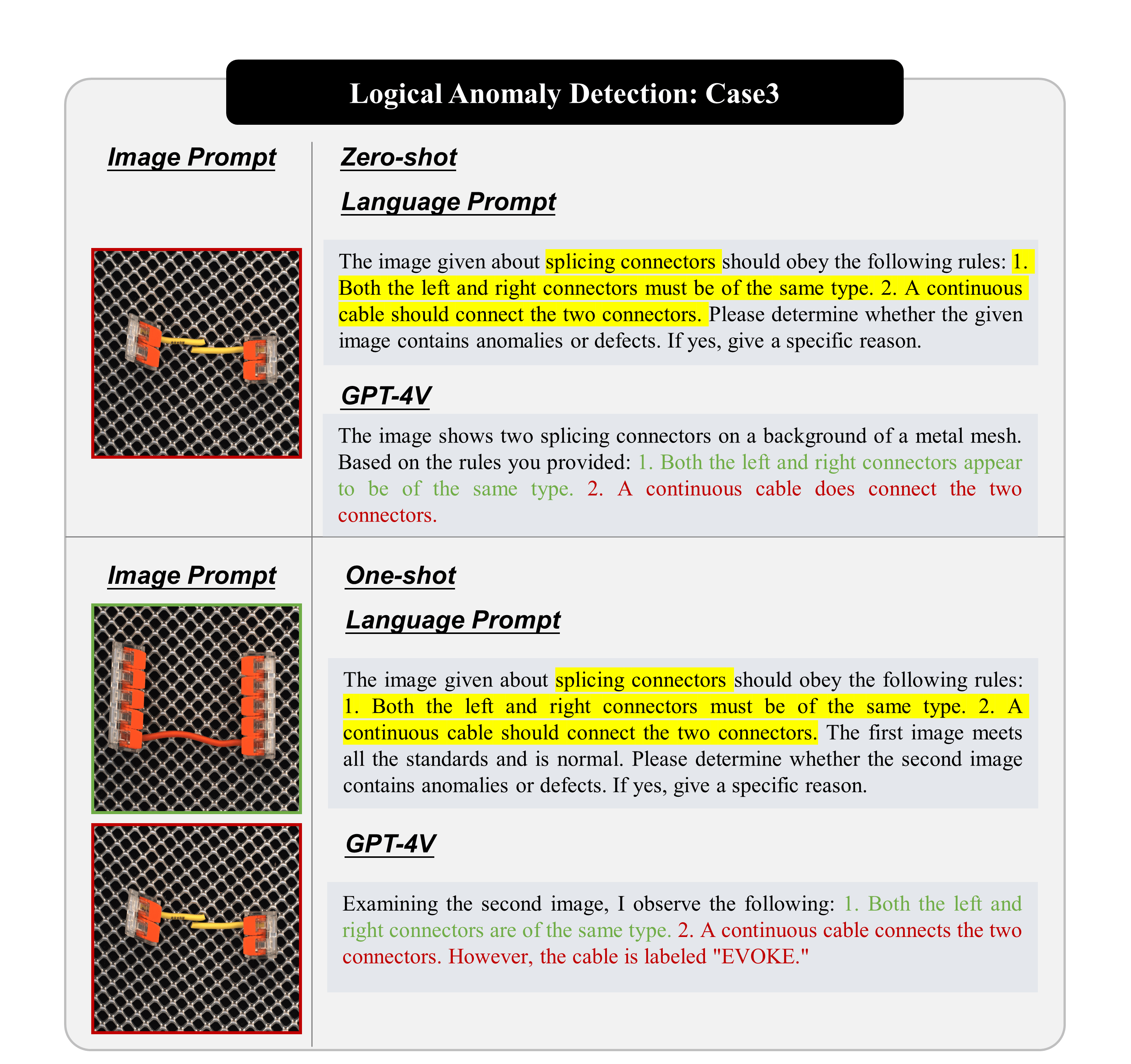}
    \vspace{3pt}
    \caption[Logical Anomaly Detection: Case 3]
    {\textbf{Logical Anomaly Detection: Case 3, the Splicing Connector category of MVTec LOCO~\cite{MVTec-LOCO}}. \hl{Yellow} highlights the given class information and normal and abnormal state descriptions.   \textcolor{HLGreen}{Green}, \textcolor{HLRed}{red}, and \textcolor{HLBlue}{blue} highlight the expected, incorrect, and additional information outputted by GPT-4V. 
}
    \vspace{3pt}
    \label{fig:logical-ad-case3}
\end{figure}

\begin{figure}[hbt!]
    \centering
    \includegraphics[width = \textwidth]{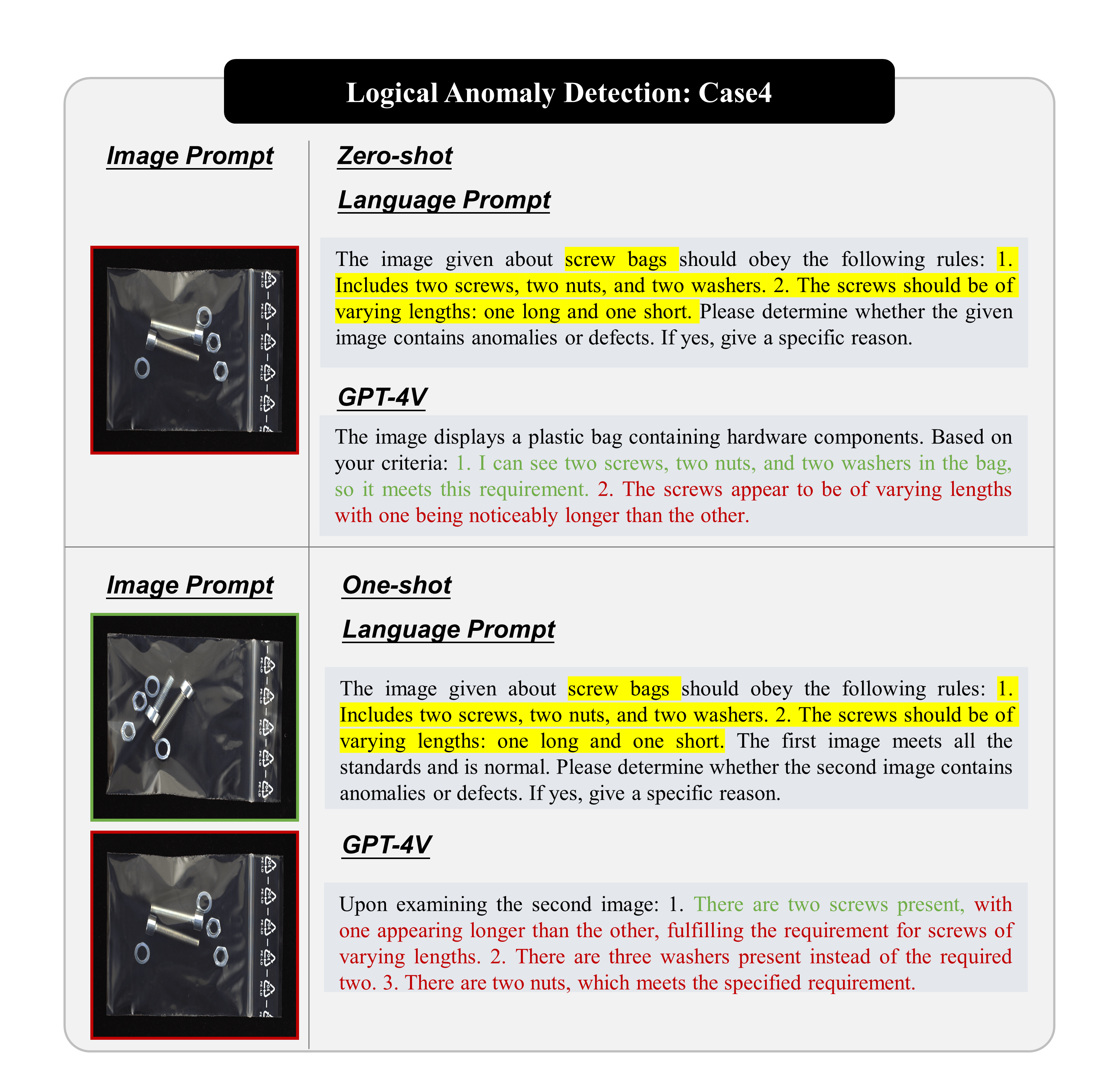}
    \vspace{3pt}
    \caption[Logical Anomaly Detection: Case 4]
    {\textbf{Logical Anomaly Detection: Case 4, the Screw Bag category of MVTec LOCO~\cite{MVTec-LOCO}}. \hl{Yellow} highlights the given class information and normal and abnormal state descriptions.   \textcolor{HLGreen}{Green}, \textcolor{HLRed}{red}, and \textcolor{HLBlue}{blue} highlight the expected, incorrect, and additional information outputted by GPT-4V. 
}
    \vspace{3pt}
    \label{fig:logical-ad-case4}
\end{figure}

The evaluation results, as depicted in Fig.~\ref{fig:logical-ad-case1},~\ref{fig:logical-ad-case2},~\ref{fig:logical-ad-case3},~\ref{fig:logical-ad-case4}, unequivocally highlight the robust image comprehension and logical reasoning capabilities of GPT-4V. For instance, in Fig.~\ref{fig:logical-ad-case1}, GPT-4V demonstrates its proficiency in interpreting intricate standards, encompassing criteria such as the presence of "1. It should contain two oranges, one peach, and some cereal, nuts, and banana slices; 2. The fruit should be on the left side of the lunch box, the cereal on the upper right, and the nuts and banana slices on the lower right of the lunch box". GPT-4V adeptly breaks down this complex task into subcomponents, identifying and localizing the various items before calculating their quantities and positions. Ultimately, GPT-4V accurately concludes that the provided breakfast box does not adhere to the stipulated standards.

Moreover, visual references play a pivotal role in enhancing GPT-4V's performance. In Fig.~\ref{fig:logical-ad-case2}, without the aid of a visual reference, GPT-4V erroneously classifies the juice bottle as a normal one. However, when presented with a referenced image, GPT-4V effectively comprehends the rule "2. To prevent bottle explosions, ensure the juice is filled to about 3cm below the bottle's opening" and delivers a correct analysis.

Nonetheless, GPT-4V may encounter challenges in scenarios where its ability to contextualize images is constrained. Notably, GPT-4V fails to detect a broken cable in Fig.~\ref{fig:logical-ad-case3} and inaccurately quantifies washers in Fig.~\ref{fig:logical-ad-case4}. The limitations of GPT-4V, particularly in matters of fine-grained details like counting, have been addressed in prior research~\cite{Dawn-of-LMMs}. Furthermore, it is worth noting that multi-round conversations and specific language prompts can significantly impact GPT-4V's performance in such cases.

\section{Medical Image Anomaly Detection}
\label{Medical Image Anomaly Detection}

\subsection{Task Introduction}
Anomaly detection, also known as outlier detection, is a pivotal task in the domain of medical imaging, aimed at identifying abnormal patterns that do not conform to expected behavior\cite{fernando2021deep}. These abnormalities or anomalies could be indicative of a wide range of medical conditions or diseases[citation]. The primary goal of anomaly detection is to accurately discern these irregularities from a plethora of medical imaging data, thereby aiding in early diagnosis and effective treatment planning. Current medical anomaly detection methods can be categorized into reconstruction-based methods~\cite{chen2020unsupervised}~\cite{gong2019memorizing}~\cite{venkataramanan2020attention}, GAN-based~\cite{2022LiuDA}, self-supervised methods~\cite{sohn2020learning} ~\cite{tian2021constrained}~\cite{tian2023self} and pre-train methods~\cite{reiss2021panda}~\cite{defard2021padim}~\cite{2023liuST}~\cite{2023liuST1} Although these methods have achieved great improvements, a unified anomaly detection model across different diseases and modalities still remains an unsolved challenge. As highlighted in ~\cite{gpt4v} and ~\cite{GPT-4V-medical}, GPT-4V, equipped with numerous multi-modal knowledge, shows promising future in enhancing the performance of anomaly detection tasks in various medical imaging modalities.

\subsection{Testing philosophy}
We aim to investigate the generalization abilities of GPT-4V on medical anomaly detection. Thus medical images on across different diseases and modalities are used, including Head MRI, Head CT, Retinal OCT, Chest X-ray and so on. For the text prompt, we also take the previous multi-step prompt to test its zero-shot and one-shot abilities. There are generally three types of prompts, a)general medical information, the disease and modalities of the medical images, such as "Chest X-ray Image" or "Head CT Image" b)human expertise, based on the general medical information, we further give the possible disease name in the medical image, e.g."The image should be classified as normal or hemorrhage", c) reference image: normal reference image to provide GPT-4V a better understanding of nomrality.

We propose to evaluate GPT-4V in either a zero-shot setting, with only language prompts, or a one-shot setting, with one reference image provided along with the language prompts. For each setting, we test three different variants: a) a naive prompt like "Please determine whether the image contains anomalies" b)general medical information, and c) with human expertise.

\subsection{Case Demonstration}
\begin{figure}[hbt!]
    \centering
    \includegraphics[width = \textwidth]{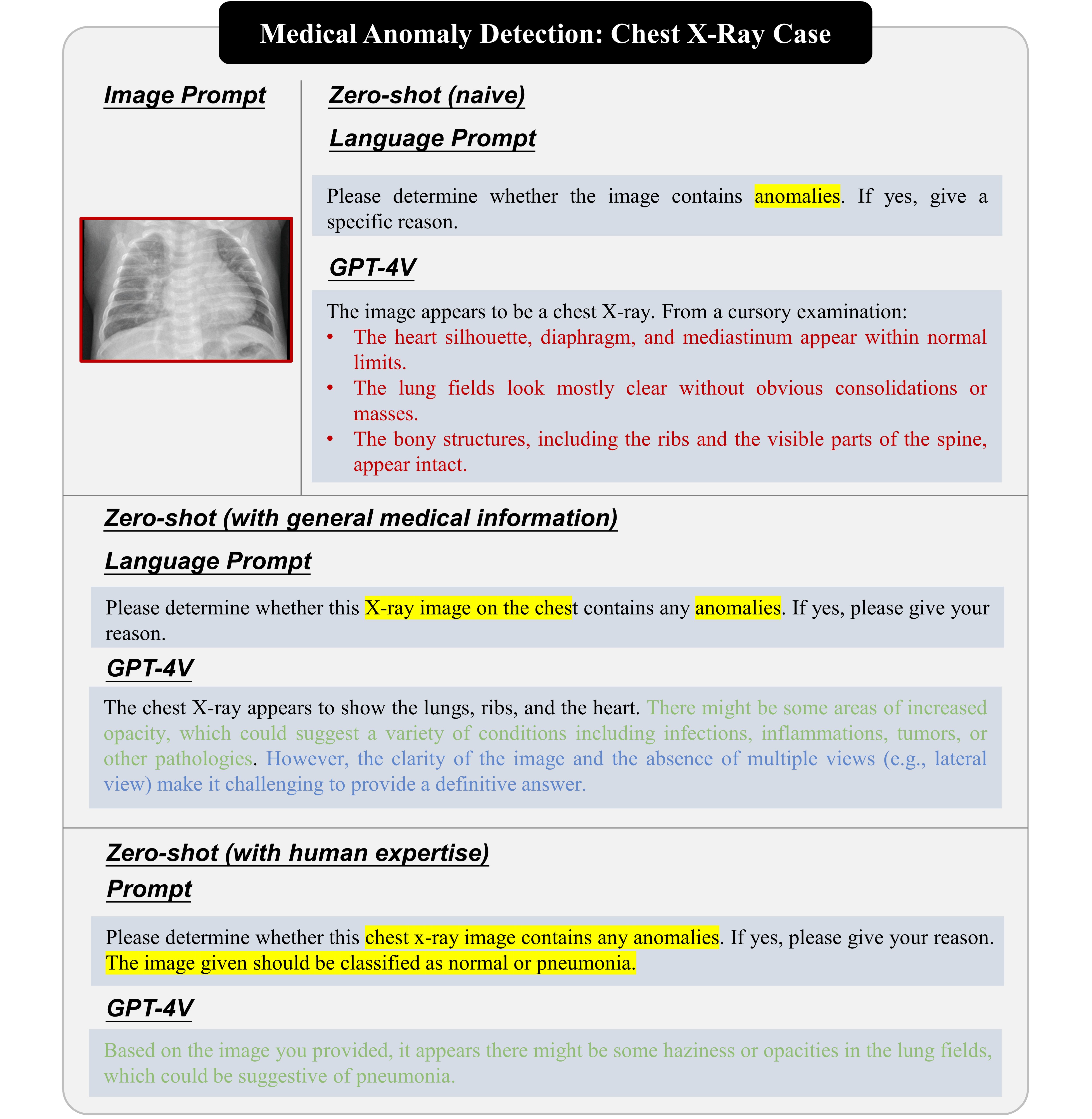}
    \vspace{3pt}
    \caption[Medical Anomaly Detection: Case 1]
    {\textbf{Medical Anomaly Detection: Case 1, the Chest X-ray~\cite{kermany2018identifying}}. \hl{Yellow} highlights the given class information and normal and abnormal state descriptions.   \textcolor{HLGreen}{Green}, \textcolor{HLRed}{red}, and \textcolor{HLBlue}{blue} highlight the expected, incorrect, and additional information outputted by GPT-4V. 
}
    \vspace{3pt}
    \label{fig:medical-ad-zeroshot1}
\end{figure}

\begin{figure}[hbt!]
    \centering
    \includegraphics[width = \textwidth]{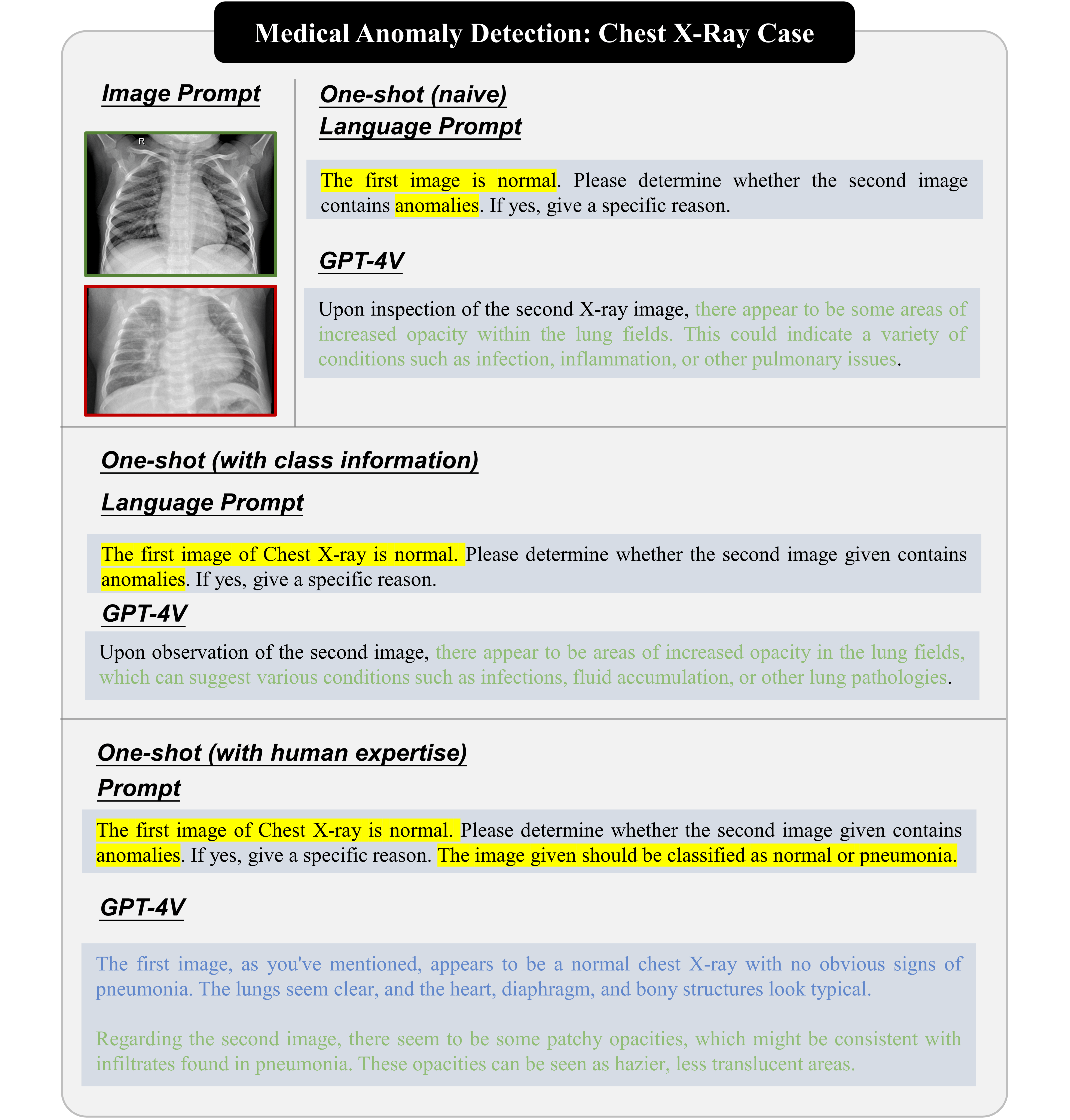}
    \vspace{3pt}
    \caption[Medical Anomaly Detection: Case 1]
    {\textbf{Medical Anomaly Detection: Case 1, the Chest X-ray~\cite{kermany2018identifying}}. \hl{Yellow} highlights the given class information and normal and abnormal state descriptions.   \textcolor{HLGreen}{Green}, \textcolor{HLRed}{red}, and \textcolor{HLBlue}{blue} highlight the expected, incorrect, and additional information outputted by GPT-4V. 
}
    \vspace{3pt}
    \label{fig:medical-ad-oneshot1}
\end{figure}

\begin{figure}[hbt!]
    \centering
    \includegraphics[width = \textwidth]{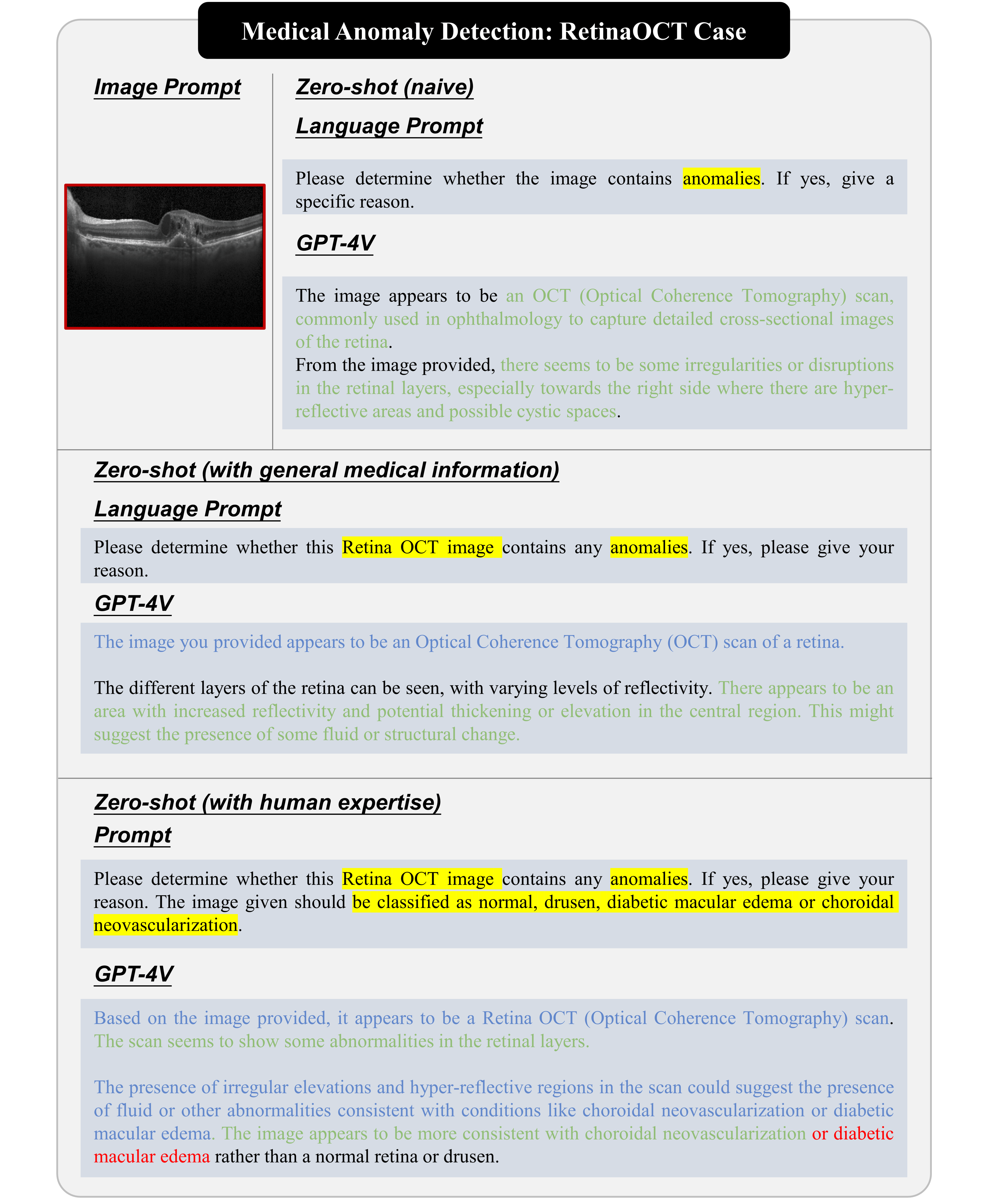}
    \vspace{3pt}
    \caption[Medical Anomaly Detection: Case 2]
    {\textbf{Medical Anomaly Detection: Case 2, the Retinal OCT~\cite{kermany2018identifying}}. \hl{Yellow} highlights the given class information and normal and abnormal state descriptions.   \textcolor{HLGreen}{Green}, \textcolor{HLRed}{red}, and \textcolor{HLBlue}{blue} highlight the expected, incorrect, and additional information outputted by GPT-4V. 
}
    \vspace{3pt}
    \label{fig:medical-ad-zeroshot2}
\end{figure}

\begin{figure}[hbt!]
    \centering
    \includegraphics[width = \textwidth]{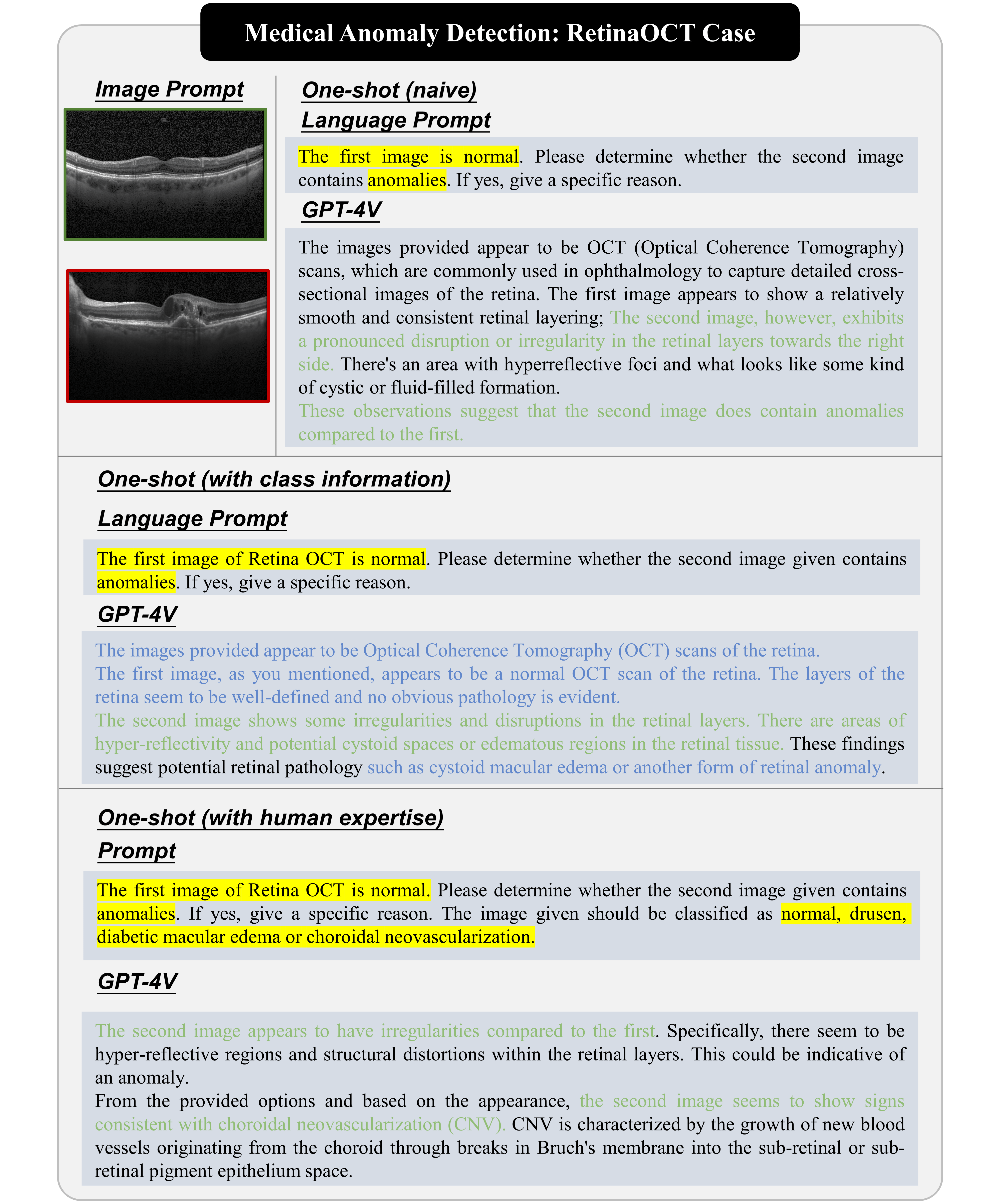}
    \vspace{3pt}
    \caption[Medical Anomaly Detection: Case 2]
    {\textbf{Medical Anomaly Detection: Case 2, the Retinal OCT~\cite{kermany2018identifying}}. \hl{Yellow} highlights the given class information and normal and abnormal state descriptions.   \textcolor{HLGreen}{Green}, \textcolor{HLRed}{red}, and \textcolor{HLBlue}{blue} highlight the expected, incorrect, and additional information outputted by GPT-4V. 
}
    \vspace{3pt}
    \label{fig:medical-ad-oneshot2}
\end{figure}

 \begin{figure}[hbt!]
     \centering
     \includegraphics[width = \textwidth]{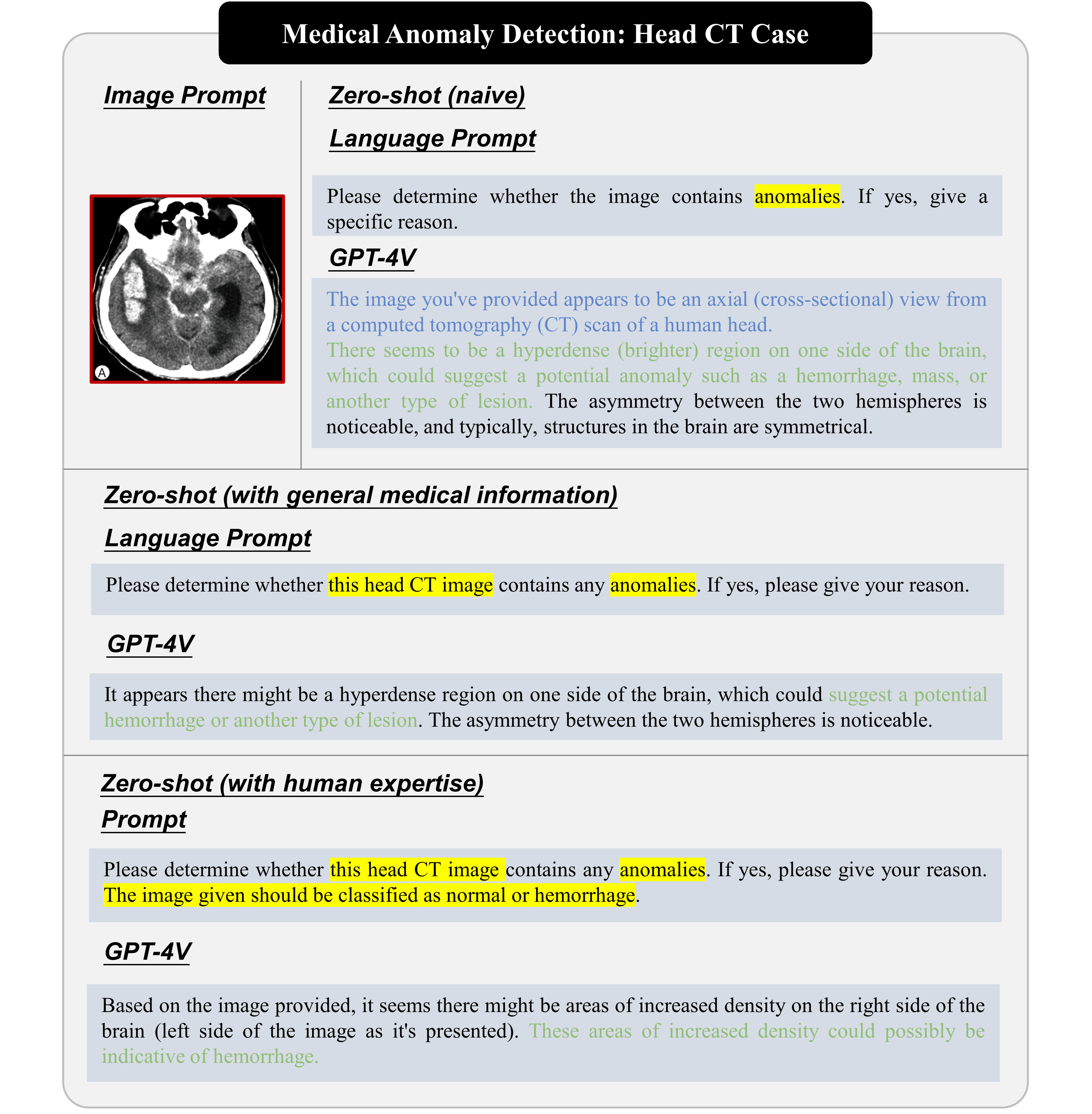}
     \vspace{3pt}
     \caption[Medical Anomaly Detection: Case 3]
     {\textbf{Medical Anomaly Detection: Case 3, the Head CT~\cite{kitamura_2018}}. \hl{Yellow} highlights the given class information and normal and abnormal state descriptions.   \textcolor{HLGreen}{Green}, \textcolor{HLRed}{red}, and \textcolor{HLBlue}{blue} highlight the expected, incorrect, and additional information outputted by GPT-4V. 
 }
     \vspace{3pt}
     \label{fig:medical-ad-zeroshot3}
 \end{figure}

\begin{figure}[hbt!]
    \centering
    \includegraphics[width = \textwidth]{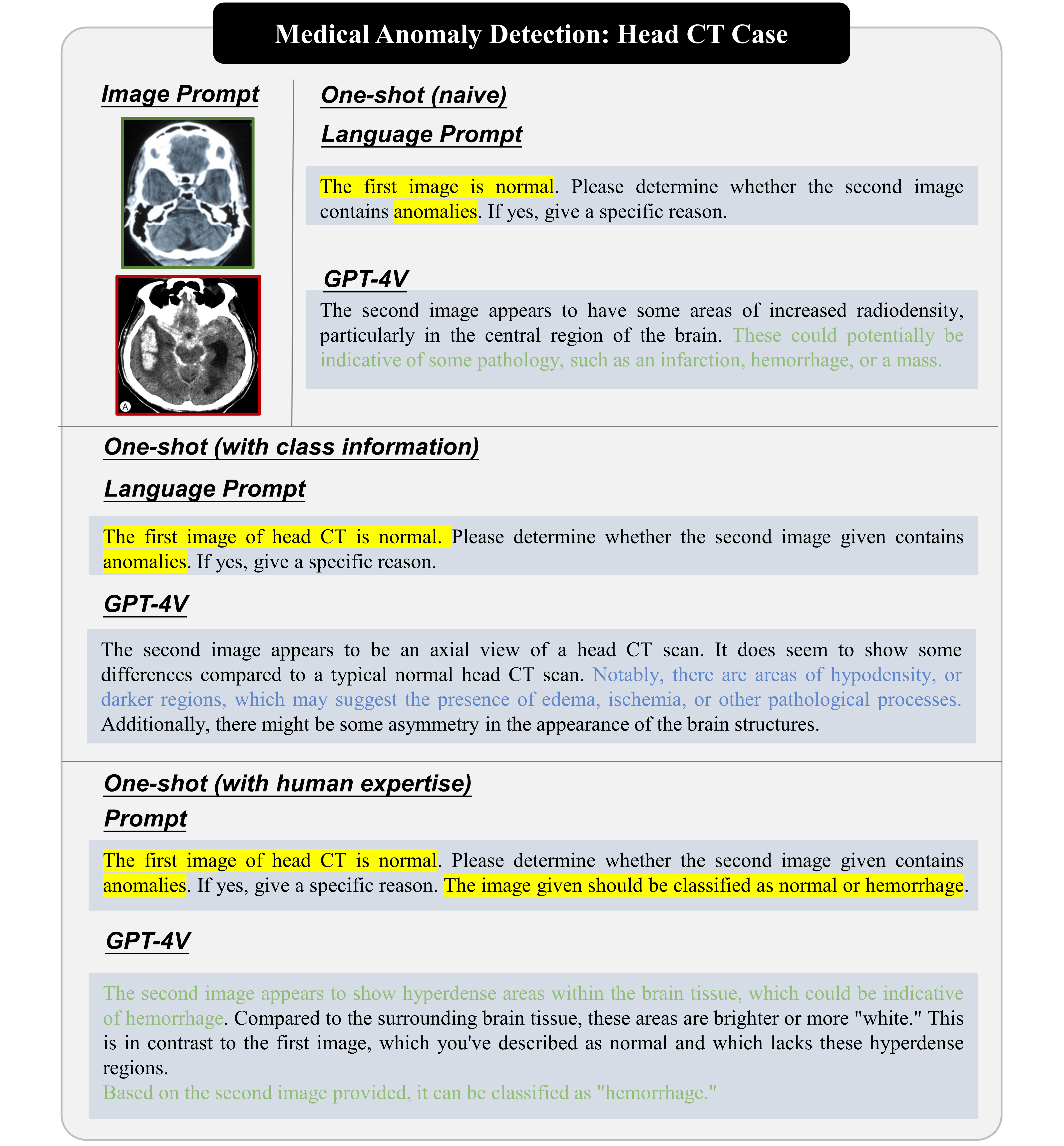}
    \vspace{3pt}
    \caption[Medical Anomaly Detection: Case 3]
    {\textbf{Medical Anomaly Detection: Case 3, the Head CT~\cite{kitamura_2018}}. \hl{Yellow} highlights the given class information and normal and abnormal state descriptions.   \textcolor{HLGreen}{Green}, \textcolor{HLRed}{red}, and \textcolor{HLBlue}{blue} highlight the expected, incorrect, and additional information outputted by GPT-4V. 
}
    \vspace{3pt}
    \label{fig:medical-ad-oneshot3}
\end{figure}

\begin{figure}[hbt!]
    \centering
    \includegraphics[width = \textwidth]{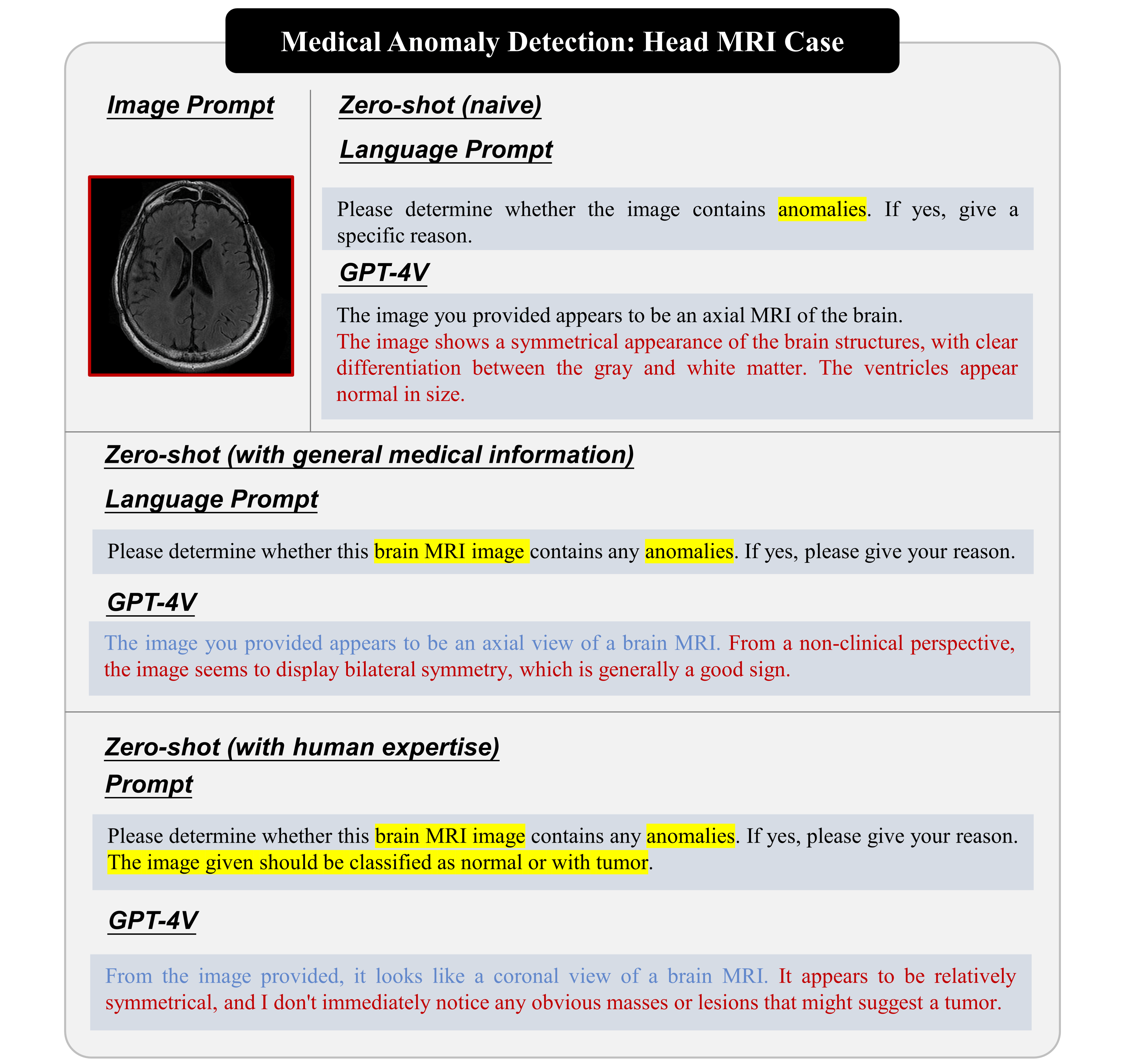}
    \vspace{3pt}
    \caption[Medical Anomaly Detection: Case 4]
    {\textbf{Medical Anomaly Detection: Case 4, Head MRI Image~\cite{Chakrabarty_2019}}. \hl{Yellow} highlights the given class information and normal and abnormal state descriptions.   \textcolor{HLGreen}{Green}, \textcolor{HLRed}{red}, and \textcolor{HLBlue}{blue} highlight the expected, incorrect, and additional information outputted by GPT-4V. 
}
    \vspace{3pt}
    \label{fig:medical-ad-zeroshot4}
\end{figure}

\begin{figure}[hbt!]
    \centering
    \includegraphics[width = \textwidth]{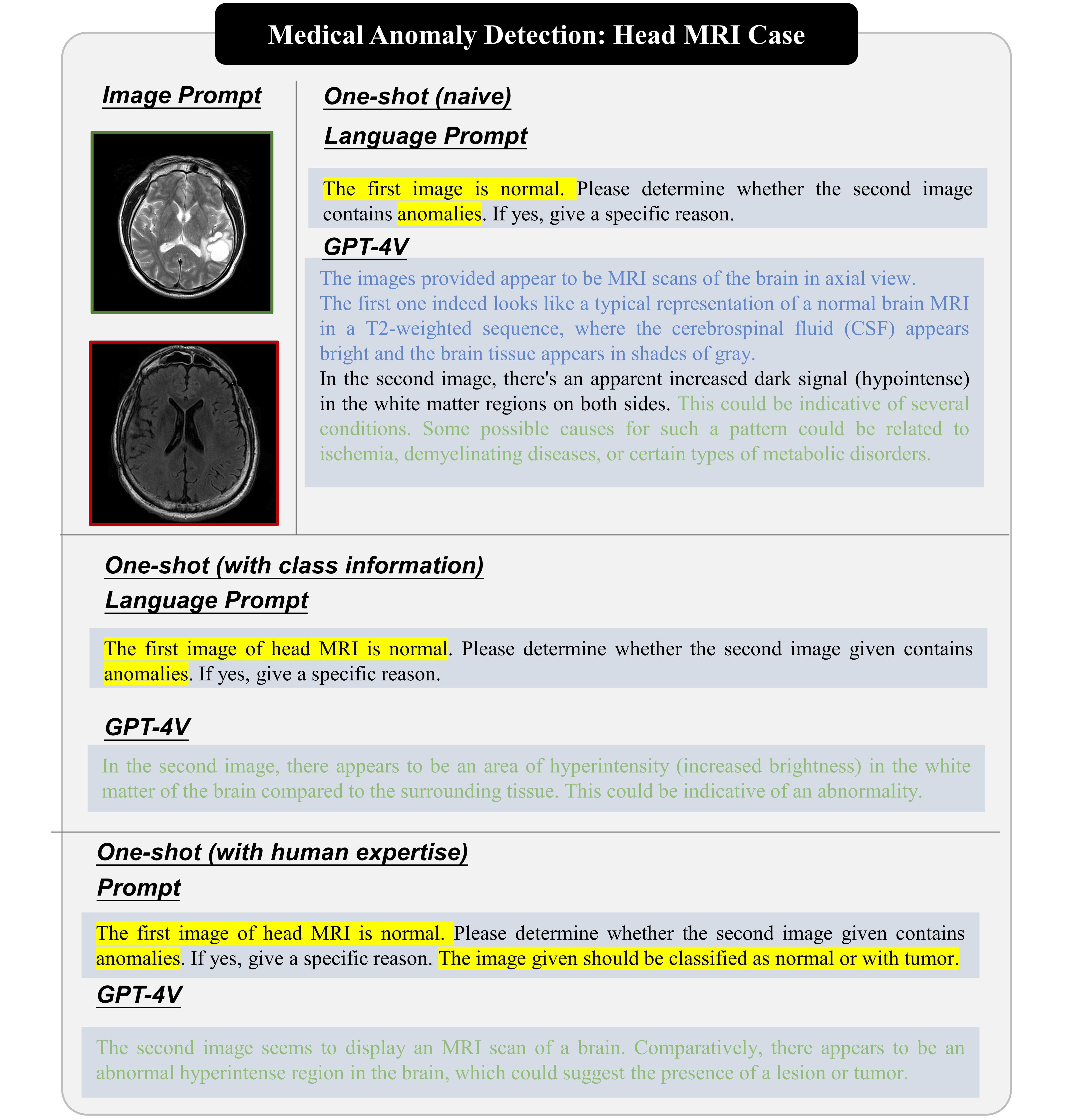}
    \vspace{3pt}
    \caption[Medical Anomaly Detection: Case 4]
    {\textbf{Medical Anomaly Detection: Case 4, Head MRI Image~\cite{Chakrabarty_2019}}. \hl{Yellow} highlights the given class information and normal and abnormal state descriptions.   \textcolor{HLGreen}{Green}, \textcolor{HLRed}{red}, and \textcolor{HLBlue}{blue} highlight the expected, incorrect, and additional information outputted by GPT-4V. 
}
    \vspace{3pt}
    \label{fig:medical-ad-oneshot4}
\end{figure}
Fig.\ref{fig:medical-ad-zeroshot1}, \ref{fig:medical-ad-zeroshot2}, \ref{fig:medical-ad-zeroshot3} and \ref{fig:medical-ad-zeroshot4} show the GPT-4V's zero-shot inference ability. GPT-4V is capable of automatically recognizing medical image modalities and anatomical structures, even without general medical information prompts. The superior image caption ability enables GPT-4V to describe the spatial and textural anomalies in the image. However, due to ethical restrictions, the GPT-4V model tends to give conservative answers when lack of sufficient information. The introduction of both general medical information and human expertise successfully leads GPT-4V to generate more concrete and accurate answers, as shown in Fig \ref{fig:medical-ad-zeroshot1}, \ref{fig:medical-ad-zeroshot2} and \ref{fig:medical-ad-zeroshot3}. However, GPT-4V fails to recognize anomalies in Fig \ref{fig:medical-ad-zeroshot4}, even with enough information provided. The abnormal area is not obvious in the image, so it turns out that it has high requirements for the medical image quality. When a visual reference is added, the GPT-4V's image caption ability successfully describe the difference between normal and abnormal images, which is shown in Fig \ref{fig:medical-ad-oneshot1}, \ref{fig:medical-ad-oneshot2} \ref{fig:medical-ad-oneshot3} and \ref{fig:medical-ad-oneshot4}.

\section{Medical Image Anomaly Localization}
\label{Medical Image Anomaly Localization}

\subsection{Task Introduction}
Following the detection of medical anomaly, the subsequent critical task is anomaly localization, which entails pinpointing the exact spatial location of the identified anomaly within the medical image~\cite{tian2021constrained}~\cite{yuan2023devil}. Accurate localization is imperative for clinicians to understand the extent and nature of the pathology, which in turn informs the course of clinical intervention. However, the real-world clinical scenario, such as tumor anomaly localization, is more complex, where either normal or abmoral cases have multiple types of tumors. Establishing a direct relationship between image pixels and excessive semantics (types of tumors) is diffcult for real world medical image anomaly localization. Several methods, including self-supervised based method~\cite{tian2021constrained} and cluster-based method~\cite{yuan2023devil} have been proposed to deal with the medical image anomaly localization task. Inspired by~\cite{SoM}, we would like to examine the localization ability of GPT-4V model, under the visual prompts. 

\subsection{Testing philosophy}
To test the GPT-4V's ability on medical image localization, we utilize several diseases categories and modalities, including abdominal CT image, endoscopy image, head MRI image and skin lesion image. Both diseased area and manually synthetic abnormal are taken into consideration to test its robustness. The visual prompts proposed by ~\cite{SoM} are also used to harness the fine-grained localization abilities of GPT-4V, including a set of image-mask pairs and corresponding index numbers to each mask. Thus, the input images are the raw images with the augmented one with masks and numbers. We also adopt a straightforward text prompt to introduce the relationship between the two input images, as follows: "The first image
needs to be inspected. The second image contains its corresponding marks. Please determine whether the image contains anomalies or defects. If yes, give a specific reason"

\subsection{Case Demonstration}
\begin{figure}[hbt!]
    \centering
    \includegraphics[width = \textwidth]{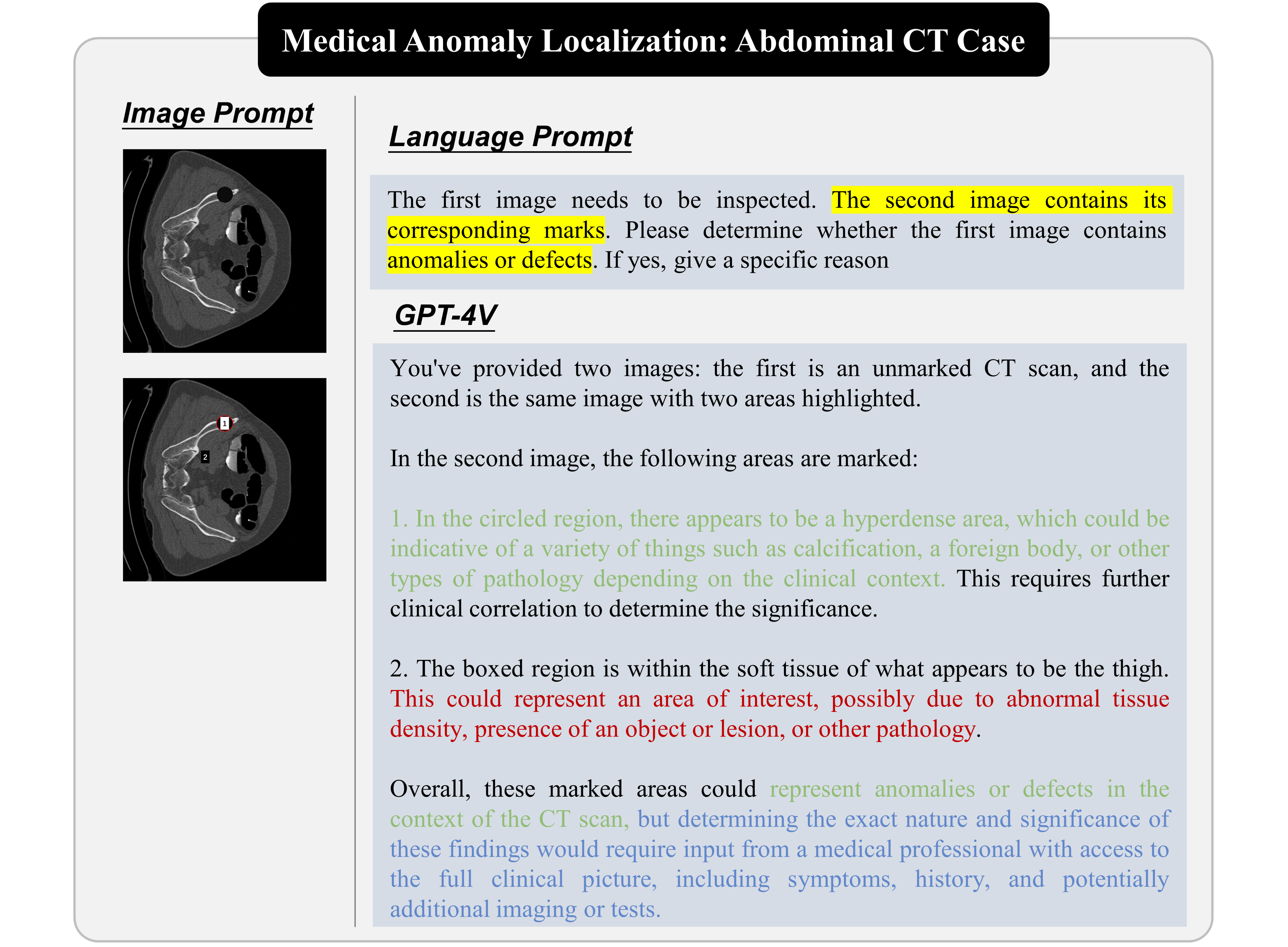}
    \vspace{3pt}
    \caption[Medical Anomaly Localization: Case 1]
    {\textbf{Medical Anomaly Localization: Case 1, Abdonimal CT Localization~\cite{9762702}}. \hl{Yellow} highlights the given class information and normal and abnormal state descriptions.   \textcolor{HLGreen}{Green}, \textcolor{HLRed}{red}, and \textcolor{HLBlue}{blue} highlight the expected, incorrect, and additional information outputted by GPT-4V. 
}
    \vspace{3pt}
    \label{fig:medical-al-case1}
\end{figure}

\begin{figure}[hbt!]
    \centering
    \includegraphics[width = \textwidth]{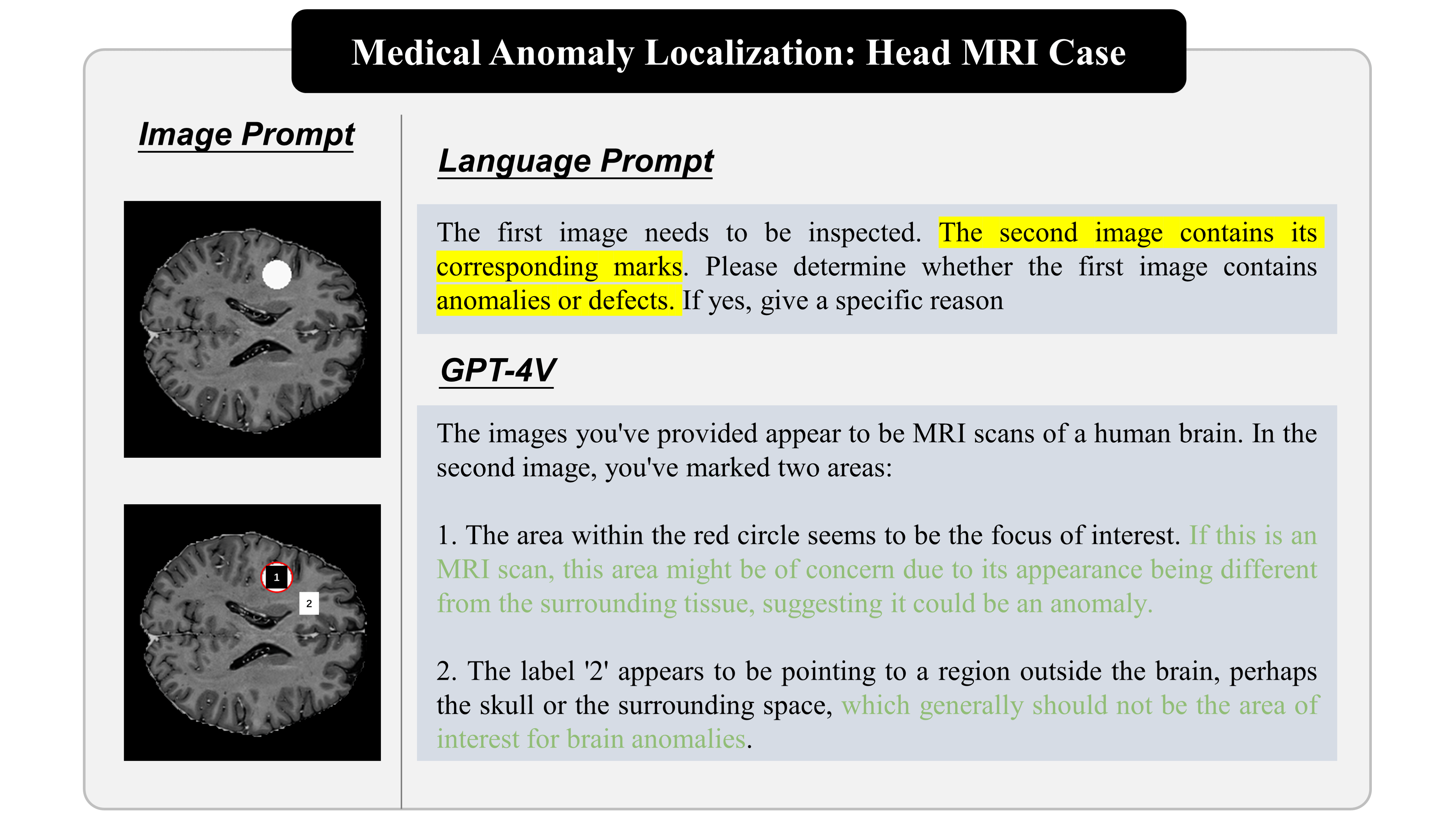}
    \vspace{3pt}
    \caption[Medical Anomaly Localization: Case 2]
    {\textbf{Medical Anomaly Localization: Case 2,Head MRI Localization~\cite{9762702}}. \hl{Yellow} highlights the given class information and normal and abnormal state descriptions.   \textcolor{HLGreen}{Green}, \textcolor{HLRed}{red}, and \textcolor{HLBlue}{blue} highlight the expected, incorrect, and additional information outputted by GPT-4V. 
}
    \vspace{3pt}
    \label{fig:medical-al-case2}
\end{figure}

\begin{figure}[hbt!]
    \centering
    \includegraphics[width = \textwidth]{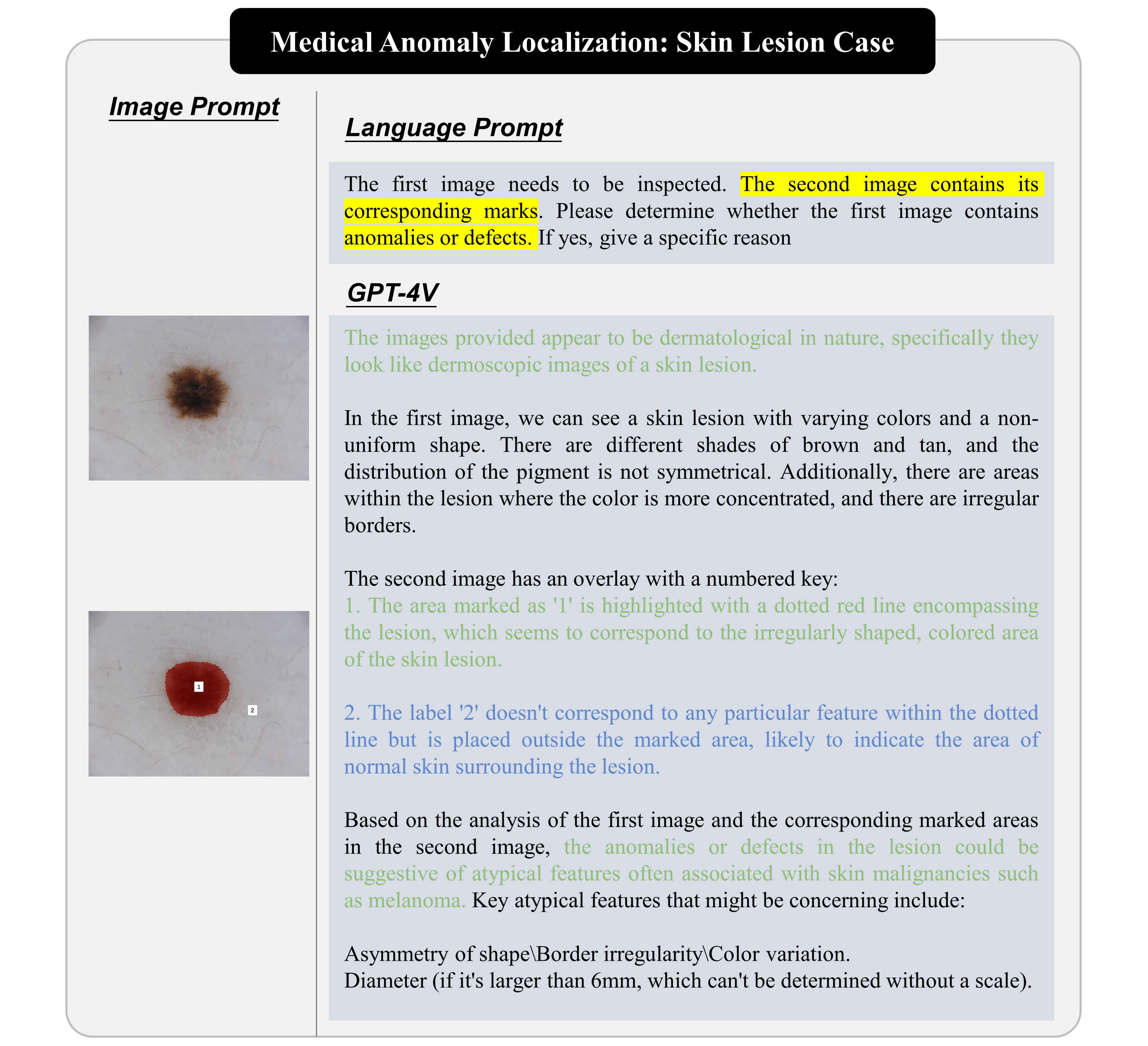}
    \vspace{3pt}
    \caption[Medical Anomaly Localization: Case 3]
    {\textbf{Medical Anomaly Localization: Case 3, Skin Lesion Localization~\cite{codella2018skin}}. \hl{Yellow} highlights the given class information and normal and abnormal state descriptions.   \textcolor{HLGreen}{Green}, \textcolor{HLRed}{red}, and \textcolor{HLBlue}{blue} highlight the expected, incorrect, and additional information outputted by GPT-4V. 
}
    \vspace{3pt}
    \label{fig:medical-al-case3}
\end{figure}

\begin{figure}[hbt!]
    \centering
    \includegraphics[width = \textwidth]{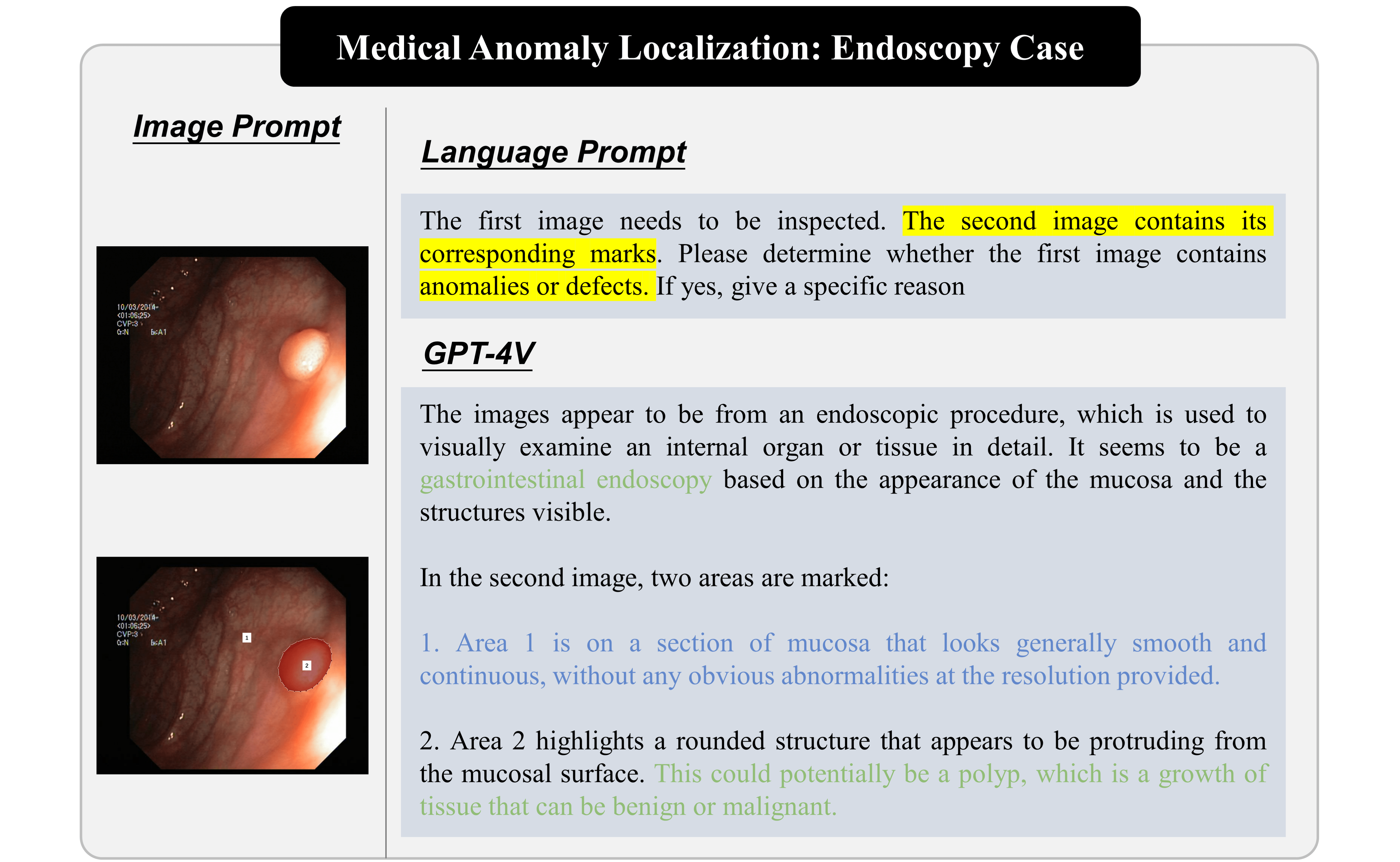}
    \vspace{3pt}
    \caption[Medical Anomaly Localization: Case 4]
    {\textbf{Medical Anomaly Localization: Case 4, Endoscopy Localization~\cite{borgli2020hyperkvasir}}. \hl{Yellow} highlights the given class information and normal and abnormal state descriptions.   \textcolor{HLGreen}{Green}, \textcolor{HLRed}{red}, and \textcolor{HLBlue}{blue} highlight the expected, incorrect, and additional information outputted by GPT-4V. 
}
    \vspace{3pt}
    \label{fig:medical-al-case4}
\end{figure}
The qualitative results are shown in Fig \ref{fig:medical-al-case1} \ref{fig:medical-al-case2} \ref{fig:medical-al-case3} and \ref{fig:medical-al-case4}. Under the instruction of visual prompts in the images, the GPT-4V tends to learn and caption the areas around the marks. For easily recognized and located cases, such as Fig \ref{fig:medical-al-case2} \ref{fig:medical-al-case3} and \ref{fig:medical-al-case4}, GPT-4V can clearly tell the difference between the anomaly areas and backgrounds. But GPT-4V fails in Fig \ref{fig:medical-al-case1}, a synthetic case where the region-of-interest shares a similar texture and shape with the background. This indicates that this model still needs to improve its detection and localization abilities under adversarial attack and complex backgrounds.

\begin{figure}[hbt!]
    \centering
    \includegraphics[width = \textwidth]{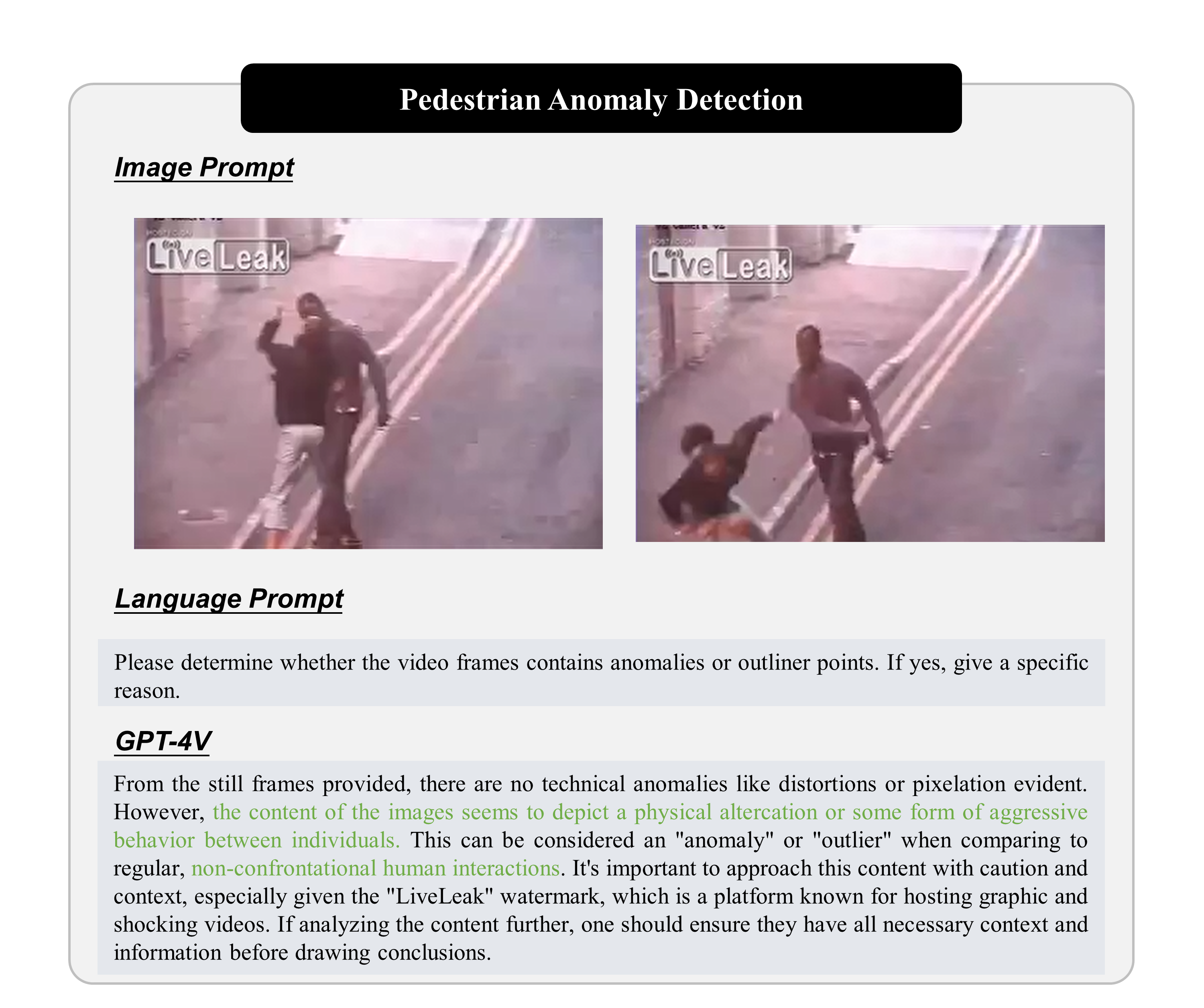}
    \vspace{3pt}
    \caption[Pedestrian Anomaly Detection]
    {\textbf{Pedestrian Anomaly Detection: Case 1, from UCF-Crime Dataset~\cite{sultani2018real}}. \textcolor{HLGreen}{Green} highlights the expected information outputted by GPT-4V. 
}
    \vspace{3pt}
    \label{fig:pedestrain-ad-case1}
\end{figure}

\section{Pedestrian Anomaly Detection}
\label{Pedestrian Anomaly Detection}

\subsection{Task Introduction}
Pedestrian anomaly detection, a subset of video anomaly detection, is dedicated to recognizing irregular activities within pedestrian interactions captured in video streams. Traditional methodologies, as referenced by various studies~\cite{amit2008,vijay2010,fouzi2015enhanced,bin2011,cewu2013,tan2016fast,zaharescu2010anomalous,ide2016sparse}, primarily rely on rule-based approaches and manually engineered features. In recent times, there has been a noticeable shift towards the adoption of deep learning techniques~\cite{hasan2016learning,chong2017abnormal,weixin2017,ravanbakhsh2017abnormal,park2020learning,li2021cutpaste,huang_self-supervised_2022,huang_self-supervision-augmented_2022,huang_weakly_2022} for pedestrian anomaly detection. The complexity of pedestrian anomaly detection arises from the need to accurately identify abnormal behaviors within the context of diverse and dynamic pedestrian interactions. This is further compounded by the varying environmental conditions in which these interactions take place. To ensure precise analysis, a substantial contextual understanding is essential.
While existing methods have demonstrated promising performance in pedestrian anomaly detection, it is worth considering that GPT-4V, with its advanced contextual comprehension capabilities, has the potential to significantly enhance the performance of this task.

\subsection{Testing philosophy}
We utilize the GPT-4V model, which currently only accepts image format visual input, for pedestrian anomaly detection. To prompt the model, we select two images from the video dataset. In addition to the image prompt, we include a simple text prompt asking the model to determine if the video frames contain anomalies or outlier points and provide a specific reason if so.

\subsection{Case Demonstration}
In Fig.~\ref{fig:pedestrain-ad-case1}, we illustrate a scenario (from UCF-Crime datadet~\cite{sultani2018real}) where a pedestrian aggresses another on the road. The GPT-4V model recognizes the aggressive behavior as an anomaly when compared to typical interactions. Additionally, it suggests caution due to the "LiveLeak" watermark, implying a need for further analysis with sufficient contextual information before drawing conclusions. The model's adeptness at discerning aggressive behavior, even in the absence of technical anomalies, demonstrates its potential to identify social anomalies within visual data.

\section{Traffic Anomaly Detection}
\label{Traffic Anomaly Detection}

\begin{figure}[hbt!]
    \centering
    \includegraphics[width = \textwidth]{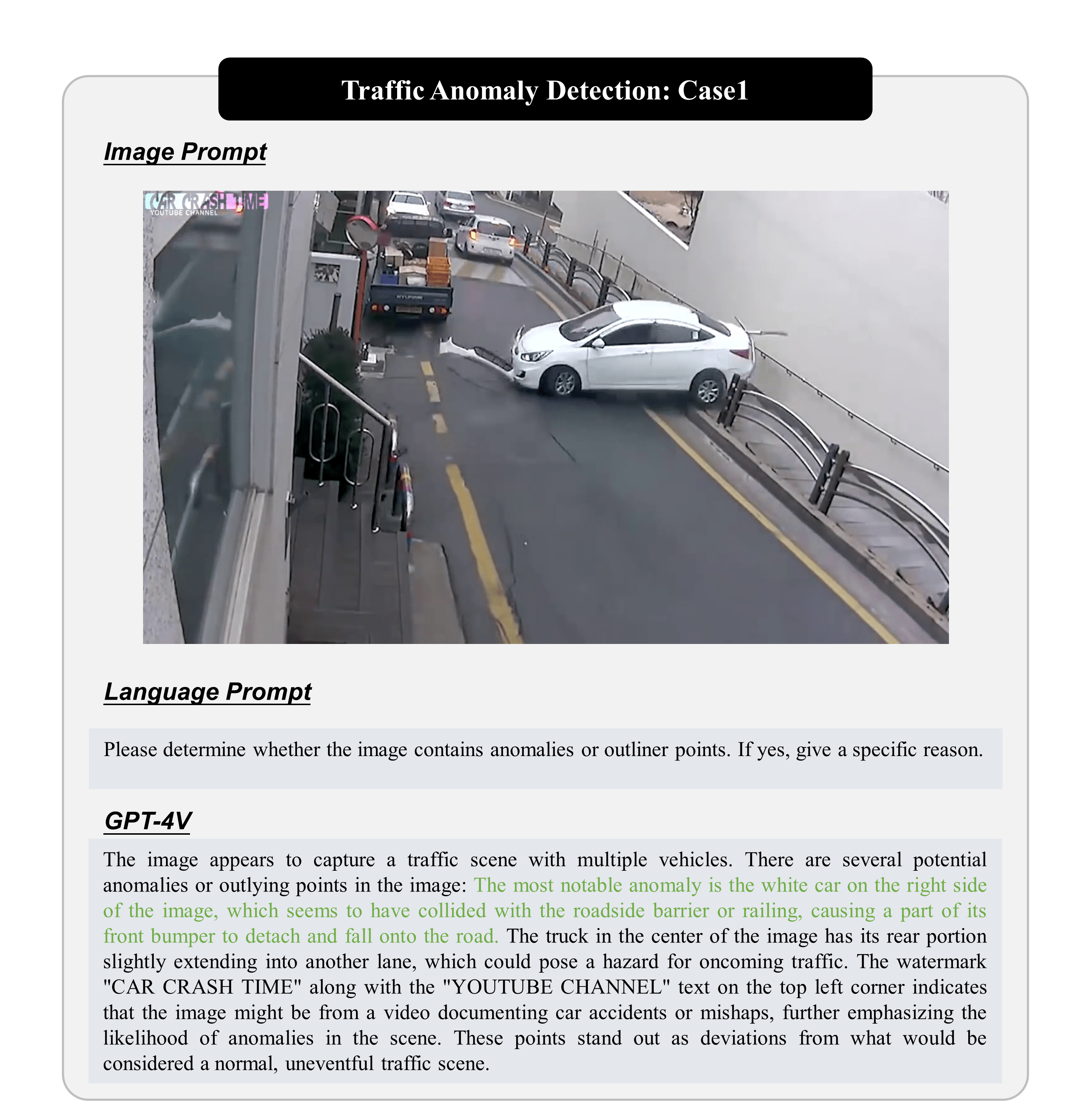}
    \vspace{3pt}
    \caption[Traffic Anomaly Detection: Case 1]
    {\textbf{Traffic Anomaly Detection: Case 1, from Kaggle Accident Detection~\cite{kaggle-accident-detection}}. \textcolor{HLGreen}{Green} highlights the expected information outputted by GPT-4V. 
}
    \vspace{3pt}
    \label{fig:traffic-ad-case1}
\end{figure}

\begin{figure}[hbt!]
    \centering
    \includegraphics[width = \textwidth]{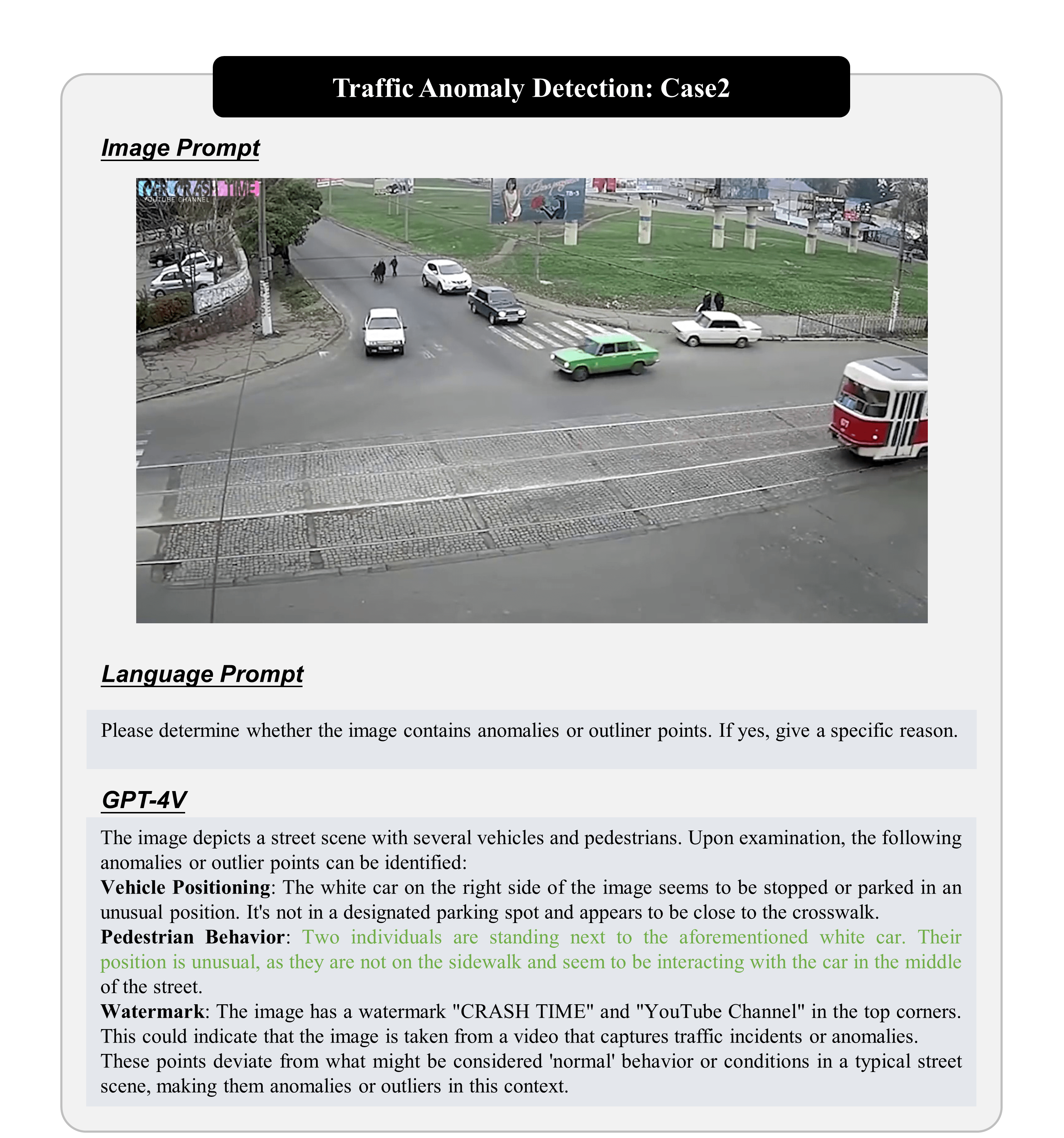}
    \vspace{3pt}
    \caption[Traffic Anomaly Detection: Case 2]
    {\textbf{Traffic Anomaly Detection: Case 2, from Kaggle Accident Detection~\cite{kaggle-accident-detection}}. \textcolor{HLGreen}{Green} highlights the expected information outputted by GPT-4V. 
}
    \vspace{3pt}
    \label{fig:traffic-ad-case2}
\end{figure}

\subsection{Task Introduction}

Traffic anomaly detection primarily aims at identifying the commencement and conclusion of abnormal events, with lesser emphasis on spatial localization. Various methodologies~\cite{hasan2016learning,medel2016anomaly,chong2017abnormal,luo2017remembering,luo2017revisit,gong2019memorizing,gong2019memorizing} have been devised to model normalcy and discern regular patterns in video frames. The prevailing challenge for anomaly detection in traffic scenarios is the development of robust algorithms that can effectively differentiate between normal and abnormal vehicles and driving behaviors, thereby ensuring the safety and reliability of the autonomous vehicle system. 
Integrating GPT4v into traffic anomaly detection promises to refine the precision and speed of current systems. GPT4v, which has the ability to conduct high-level understanding, is adept at parsing the intricacies of traffic data, thereby sharpening the discrepancy between normal variations and true anomalies. This precision is critical for developing real-time monitoring systems that deliver accurate alerts while minimizing false positives.


\subsection{Testing philosophy}
We employ GPT-4V for traffic anomaly detection, which, as of now, only accepts visual input in image format. To engage the model, we select a representative image from the traffic scene, accompanied by a succinct text prompt. This prompt requests the model to ascertain whether the image frames harbor anomalies or outlier points, and if found, to elucidate the specific reasons for such irregularities.

\subsection{Case Demonstration}
As depicted in Fig.~\ref{fig:traffic-ad-case1} and~\ref{fig:traffic-ad-case2}, by scrutinizing the spatial-temporal dynamics within the traffic scenes from a traffic anomaly detection dataset~\cite{kaggle-accident-detection}, GPT-4V proficiently differentiates between standard traffic flow and anomalous events. Beyond merely identifying outliers in traffic patterns, the model extends its utility by offering insightful elucidations concerning the abnormal nature of the scenarios. For instance, in Fig.\ref{fig:traffic-ad-case1}, the model effectively explicates an abnormal vehicular maneuver that collides with the roadside barrier and deviates from typical driving behavior. Harnessing its deep comprehension of the underlying patterns and relationships within the traffic data, the model employs interpretable techniques to unravel the factors contributing to the anomaly, thereby providing a nuanced understanding that could be pivotal for enhancing the safety and reliability of autonomous driving systems.


\begin{figure}[hbt!]
    \centering
    \includegraphics[width = \textwidth]{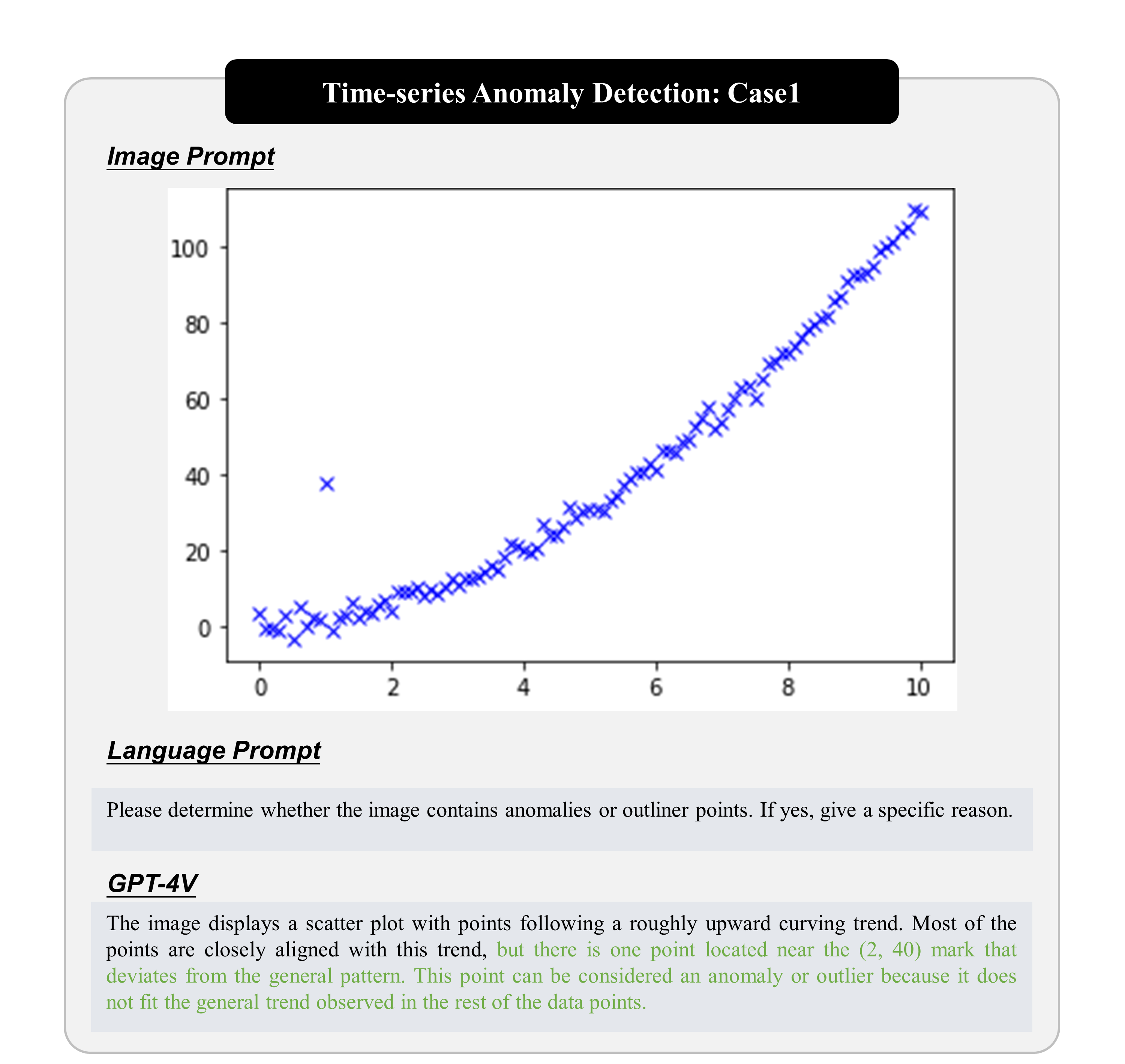}
    \vspace{3pt}
    \caption[Time Series Anomaly Detection: Case 1]
    {\textbf{Time Series Anomaly Detection: Case 1, from Outlier Detection Dataset~\cite{stackexchange-outlier-detection}.}\textcolor{HLGreen}{Green} highlights the expected information outputted by GPT-4V. 
}
    \vspace{3pt}
    \label{fig:time-series-ad-case1}
\end{figure}

\begin{figure}[hbt!]
    \centering
    \includegraphics[width = \textwidth]{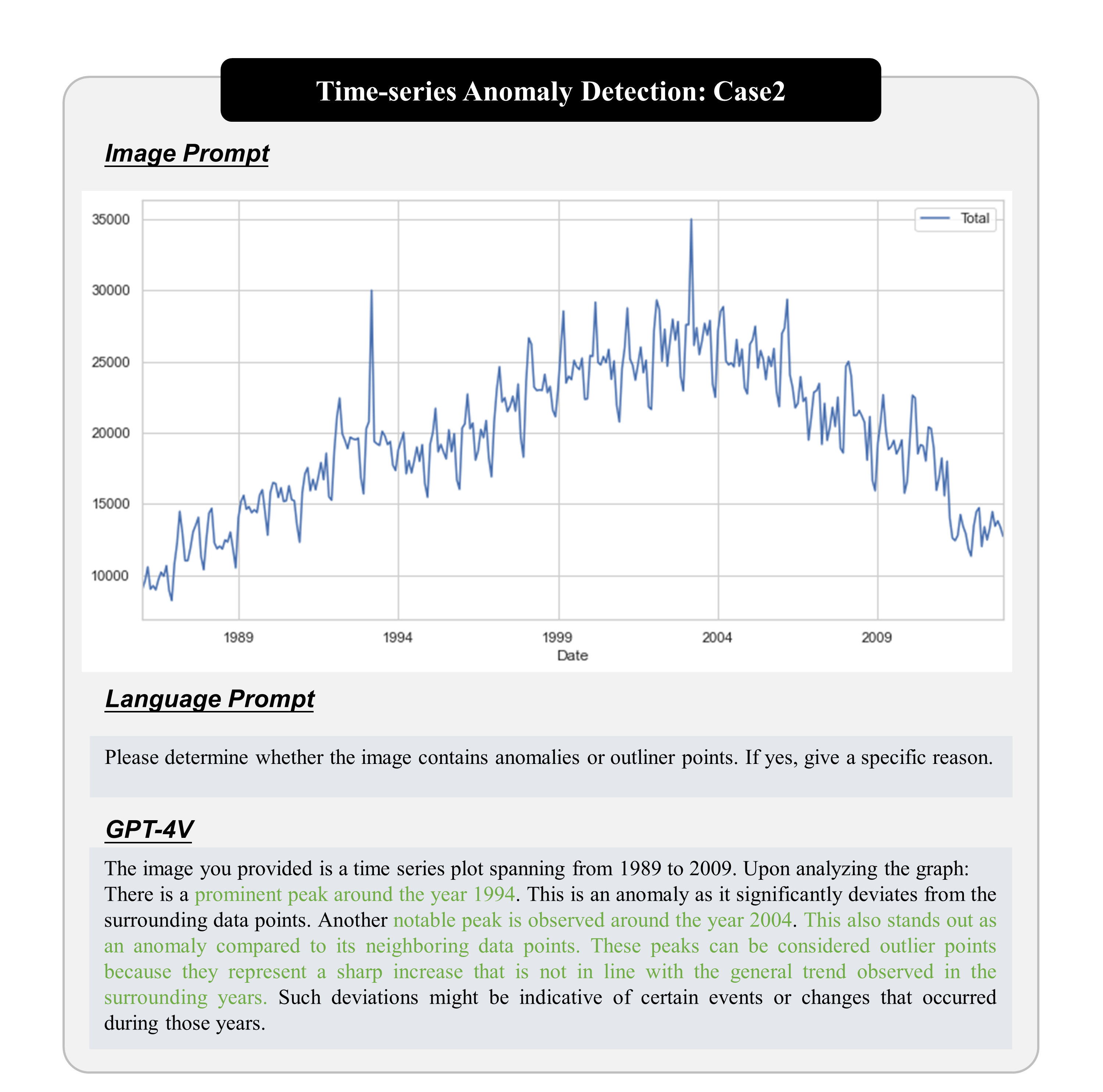}
    \vspace{3pt}
    \caption[Time Series Anomaly Detection: Case 2]
    {\textbf{Time Series Anomaly Detection: Case 2, from Catfish Sales Dataset~\cite{neptuneai-anomaly-detection}}. \textcolor{HLGreen}{Green} highlights the expected information outputted by GPT-4V. 
}
    \vspace{3pt}
    \label{fig:time-series-ad-case2}
\end{figure}

\section{Time Series Anomaly Detection}
\label{Time Series Anomaly Detection}

\subsection{Task Introduction}
Time series anomaly detection refers to the task of identifying unusual or abnormal patterns, events, or behaviors in sequential data over time, that deviate significantly from the expected or normal behavior. Time series anomaly detection models can be categorized as supervised or unsupervised algorithms. Supervised methods perform well when anomaly labels are available, such as AutoEncoder~\cite{sakurada2014anomaly} and RobustTAD\cite{gao2020robusttad}. Unsupervised algorithms are suitable when obtaining anomaly labels is challenging. This has led to the development of new unsupervised methods, including DAGMM~\cite{zong2018deep} and OmniAnomaly~\cite{su2019robust}. Unsupervised deep learning methods excel in time series anomaly detection, leveraging representation learning and a reconstruction approach to accurately identify anomalies without the need for labeled data~\cite{zhao2020multivariate, jiao2022timeautoad, zhang2022adaptive}.

\subsection{Testing philosophy}
To exploit GPT-4V for time series anomaly detection, we plot time series into images and then deliver the testing data to GPT-4V. Specifically, we select two instances ~\cite{neptuneai-anomaly-detection,stackexchange-outlier-detection} along with a simple text prompt asking the model to determine if the image contains anomalies or outlier points and provide a specific reason if so. 


\subsection{Case Demonstration}
As illustrated in Fig.~\ref{fig:time-series-ad-case1} and \ref{fig:time-series-ad-case2}, by examining the temporal dependencies and trends within the time series, GPT-4V adeptly differentiates between normal fluctuations and anomalous behavior. Beyond merely detecting outliers in the time series curves, the model extends its utility by offering insightful explanations regarding the abnormal nature of the data. For instance, in Fig.~\ref{fig:time-series-ad-case2}, the model effectively elucidates the abnormal peak in the time series. Drawing upon its profound understanding of the underlying patterns and relationships within the data, the model employs interpretability techniques to illuminate the factors contributing to the anomaly.

\section{Prospect}

The future evaluation and utilization of GPT-4V for anomaly detection hold significant promise in addressing complex challenges across various domains. As a versatile language model, GPT-4V demonstrates its potential in anomaly detection, and the following prospects aim to refine its capabilities, foster integration, and elevate its performance.

\begin{enumerate}
    \item \textbf{Quantitative Analysis}:
    Incorporating quantitative metrics, such as Precision, Recall, and F1-score, alongside AUC-ROC and MAP, in future evaluations will provide a more comprehensive understanding of GPT-4V's anomaly detection performance. This quantification will empower a more objective assessment of the model's capabilities and its adaptation to diverse anomaly detection tasks.
    
    \item \textbf{Expanding Evaluation Scope}:
    Expanding the scope to include real-world challenges, such as varying lighting conditions and occlusions in image-based anomaly detection, and different types of anomalies in time-series data, offers a more realistic view of GPT-4V's adaptability and limitations. The inclusion of synthetic and real-world anomalies adds depth to the evaluation process.

    \item \textbf{Multi-round Interaction Evaluation}:
    The potential of multi-round conversations for GPT-4V's iterative learning and adaptation to feedback provides a dynamic framework for enhancing its performance in anomaly detection. It is a promising avenue for scenarios where ongoing refinement is crucial, such as cybersecurity.
    
    \item \textbf{Incorporation of Human Feedback}:
    Utilizing human feedback loops presents the opportunity for domain experts to refine GPT-4V's understanding of complex or nuanced anomalies. The collaboration between the model and experts promises to address real-world challenges effectively.
    
    \item \textbf{Integration of Auxiliary Data}:
    Exploring the impact of integrating auxiliary data, such as additional sensor readings or metadata, is instrumental in enhancing GPT-4V's understanding and accuracy in identifying anomalies across various domains. This comprehensive approach aligns with real-world data scenarios.
    
    \item \textbf{Comparison with Specialized Models}:
     Comparative evaluations against specialized anomaly detection models are essential to identify the specific strengths and weaknesses of GPT-4V. These assessments will clarify the domains and use cases where GPT-4V's versatility excels or where specialized models remain superior.
    
    \item \textbf{Real-Time Performance Assessment}:
    Evaluating GPT-4V's real-time performance is crucial for applications requiring rapid anomaly detection. This prospect ensures the model's suitability for time-critical or online anomaly detection tasks.
    
    \item \textbf{Transfer Learning Evaluation}:
    Assessing the effectiveness of transfer learning in fine-tuning GPT-4V for specific anomaly detection tasks can pave the way for broader generalization. It enhances the model's adaptability in diverse anomaly detection scenarios.
    
    \item \textbf{Hybrid Model Development}:
    The development of hybrid models combining GPT-4V with other machine learning or deep learning approaches offers an innovative approach to address anomaly detection challenges. These hybrids aim to leverage GPT-4V's linguistic capabilities while enhancing its performance in specialized scenarios.
\end{enumerate}

In summation, these prospects set the stage for a comprehensive and multifaceted exploration of GPT-4V's anomaly detection potential. By combining quantitative metrics, real-world challenges, human feedback, auxiliary data integration, comparative assessments, and real-time capabilities, we can unlock the full scope of GPT-4V's utility in addressing anomalies across diverse fields. The journey towards improved anomaly detection with GPT-4V is one of collaboration, adaptation, and innovation, promising exciting developments in the years to come.

\section{Conclusion}

In conclusion, the assessment of GPT-4V's capabilities in anomaly detection signifies a notable advancement in the realm of versatile and adaptable AI models. GPT-4V demonstrates exceptional proficiency in identifying anomalies across diverse modalities and fields, offering both comprehensive and nuanced semantic comprehension. Its ability to deduce anomalies and its responsiveness to an expanding array of prompts underscore its versatility and potential. Nevertheless, like any technology, there remains room for further enhancement, particularly in intricate and subtle scenarios.

The opportunities delineated in this evaluation propose promising avenues for future research and development. The inclusion of quantitative metrics, broadening the spectrum of evaluations, embracing human input, and integrating supplementary data all contribute to augmenting the performance of GPT-4V. Comparative assessments against specialized models and the exploration of hybrid models further enrich the landscape of anomaly detection. Real-time assessment and the incorporation of transfer learning hold the promise of addressing time-sensitive situations and generalizing anomaly detection across diverse domains.

As we embark on this journey to unlock the full potential of GPT-4V, collaboration, adaptability, and innovation will serve as the foundational pillars of our success. The evaluation and utilization of GPT-4V for anomaly detection do not merely signify an exploration of technology but also serve as a testament to the ongoing evolution of AI and its transformative impact on real-world applications. Keeping these prospects in mind, the future of anomaly detection holds significant promise, and GPT-4V stands at the forefront of this captivating evolution.

\clearpage

\bibliographystyle{sn-mathphys} 

\end{document}